\documentclass[letterpaper, 10 pt, journal, twoside]{IEEEtran}
\usepackage{mathtools}
\usepackage{amsmath}
\usepackage{graphicx}
\usepackage{amssymb}
\usepackage{booktabs}
\usepackage{multirow}
\usepackage{multicol}
\usepackage{verbatim}
\usepackage{float}
\usepackage{subfigure}
\usepackage{kotex}
\usepackage{cite}
\usepackage{tabularx}
\usepackage{array}
\usepackage[flushleft]{threeparttable}
\usepackage[ruled,vlined]{algorithm2e}
\usepackage[dvipsnames]{xcolor}
\usepackage{dblfloatfix} 
\definecolor{bhgreen}{RGB}{0, 176, 80}
\definecolor{yeblue}{RGB}{0, 112, 192}
\definecolor{myyellow}{RGB}{255, 192, 0}
\definecolor{tabGray}{gray}{0.94}
\makeatletter
\let\NAT@parse\undefined
\makeatother
\usepackage{hyperref}
\usepackage{xcolor,colortbl}
\newcolumntype{a}{>{\columncolor{Gray}}c}
\newcolumntype{C}{>{\centering\arraybackslash}p{7em}}

\ifCLASSINFOpdf
\else
\fi
\begin{document}
\title{STEP: State Estimator for Legged Robots Using a Preintegrated Foot Velocity Factor}

% \author{Michael~Shell,~\IEEEmembership{Member,~IEEE,}
%         John~Doe,~\IEEEmembership{Fellow,~OSA,}
%         and~Jane~Doe,~\IEEEmembership{Life~Fellow,~IEEE}% <-this % stops a space
% \thanks{M. Shell was with the Department
% of Electrical and Computer Engineering, Georgia Institute of Technology, Atlanta,
% GA, 30332 USA e-mail: (see http://www.michaelshell.org/contact.html).}% <-this % stops a space
% \thanks{J. Doe and J. Doe are with Anonymous University.}% <-this % stops a space
% \thanks{Manuscript received April 19, 2005; revised August 26, 2015.}}
\author{Yeeun Kim$^{1,\star}$, Byeongho Yu$^{1,\star}$, Eungchang Mason Lee$^{1}$, Joon-ha Kim$^{2}$,\\ Hae-won Park$^{2}$, and Hyun Myung$^{*}$, \IEEEmembership{Senior~Member, IEEE \vspace{-3mm}}%
\thanks{Manuscript received: September, 9, 2021; Revised December, 18, 2021; Accepted January, 25, 2022.}%Use only for final RAL version
\thanks{This paper was recommended for publication by Editor Abderrahmane Kheddar upon evaluation of the Associate Editor and Reviewers' comments.
This work was supported partially by the Defense Challengeable Future Technology Program of Agency for Defense Development, Republic of Korea and partially by the Institute of Information $\&$ communications Technology Planning $\&$ Evaluation (IITP) grant funded by the Korea government(MSIT) (No. 2021-0-00230, development of real·virtual environmental analysis based adaptive interaction technology). The students are supported by BK21 FOUR. The Mini Cheetah robot was provided by MIT Biomimetic Robotics Lab and Naver Labs Corporation.} %Use only for final RAL version
\thanks{$\star$These authors contributed equally.}
\thanks{$^{1}$Y. Kim, B. Yu, and E. M. Lee are with School of Electrical Engineering, Korea Advanced Institute of Science and Technology~(KAIST), Daejeon, Republic of Korea
{\tt\footnotesize \{yeeunk, bhyu, eungchang$\_$mason\}@kaist.ac.kr}.}%
\thanks{$^{2}$J. Kim and H. Park are with Department of Mechanical Engineering, KAIST, Daejeon, Republic of Korea
{\tt\footnotesize \{kjhpo226, haewonpark\}@kaist.ac.kr}.}%
\thanks{$^{*}$Corresponding author: H. Myung is with School of Electrical Engineering and KI-AI, KAIST, Daejeon, Republic of Korea
{\tt\footnotesize hmyung@kaist.ac.kr}.}%
\thanks{Digital Object Identifier (DOI): see top of this page.}
}

\markboth{IEEE Robotics and Automation Letters. Preprint Version. January, 2022}
{Kim, \MakeLowercase{and} Yu \MakeLowercase{\textit{et al.}}: STEP: State Estimator for Legged Robots} 




\maketitle

\begin{abstract}
We propose a novel state estimator for legged robot\textcolor{black}{s}, \textit{STEP}, achieved through a novel {preintegrated foot velocity factor}. In the preintegrated foot velocity factor, the usual {non-slip} assumption is not adopted. Instead, the end effector velocity becomes observable by exploiting the body speed obtained from a stereo camera. In other words, the preintegrated end effector's pose can be estimated. Another advantage of our approach is that it eliminates the necessity for a contact detection step, unlike the typical approaches. The proposed method has also been validated in harsh-environment simulations and real-world experiments containing uneven or slippery terrains.
\end{abstract}

% \begin{IEEEkeywords}
% IEEE, IEEEtran, journal, \LaTeX, paper, template.
% \end{IEEEkeywords}

% \fi
\IEEEpeerreviewmaketitle

\section{Introduction}\label{sec:introduction}

\IEEEPARstart{L}{egged robots} are often needed because wheeled robots cannot navigate rough terrains and UAVs cannot carry heavy items due to its payload limitations. As the need for legged robots increases, many studies for accurate state estimation of legged robots have been conducted.
%However, {establishing} perfect {dynamic}s model {of a legged robot} {is difficult because} it depends on many {sensor information} corrupted by noise, including the joint encoder measurements and inexact kinematics \cite{3bloesch2013state}. State estimation for a legged robot is also considered challenging.

%With the recent research fusing various sensors such as a camera, an IMU and joint encoders, more robust state estimation systems are built and validated through {various} experiment{s}. These {resulted in} substantially increased availability of legged robots in various environments.

One outstanding research {for state estimation of legged robots on unstable and slippery terrain} is {a} stochastic filtering-based method\cite{3bloesch2013state}. The key contribution of {their approach} is {the introduction of} a leg kinematics constraint during {non-slip} contact. The effects coming from a possible slip are considered as a Gaussian noise. This framework is still considered a fundamental {element} of a legged robot's state estimation. 
To demonstrate the effectiveness of {their} approach, {they} use the contact sensor because the accurate contact detection is necessary for utilizing the leg kinematics constraint.

\begin{figure}[t]
    \centering
    \includegraphics[width=1.0\linewidth]{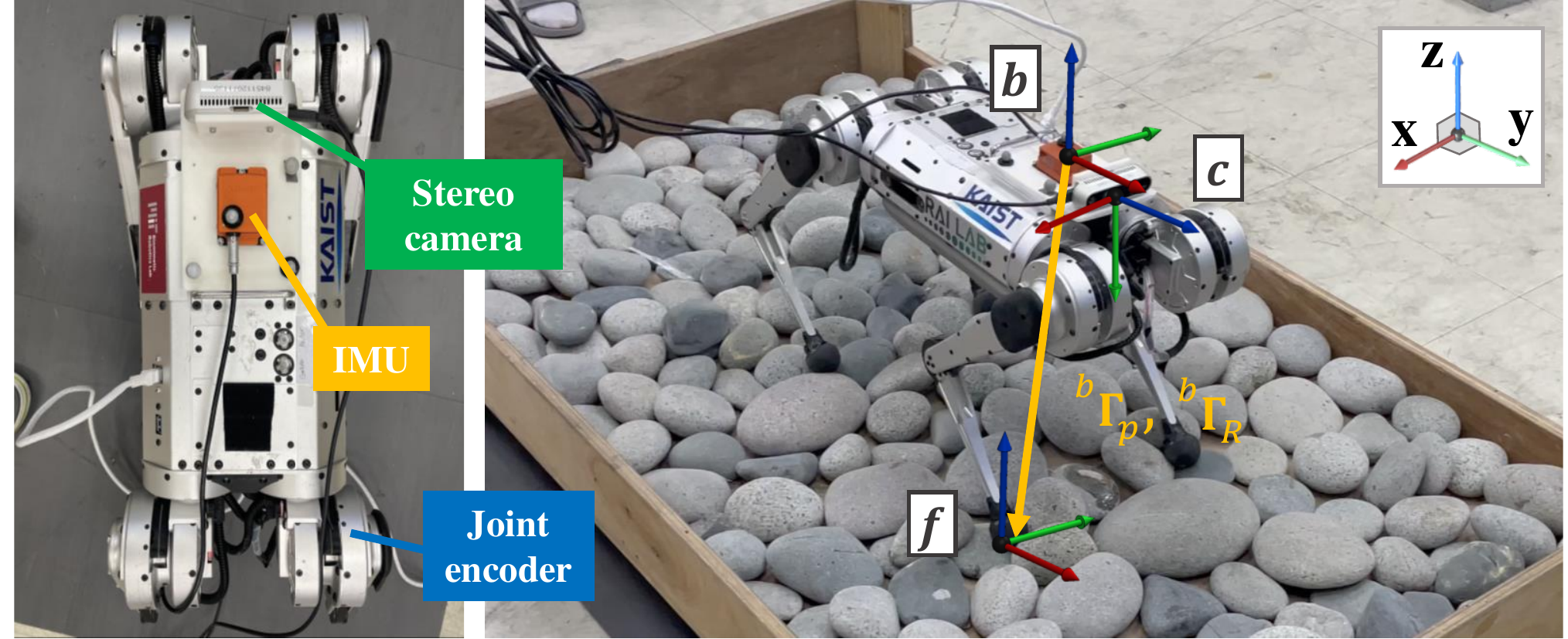}
    \caption{The legged robot \cite{6katz2019mini} and {installed} sensor setup for the real-world experiment on gravels. Each robot leg contains three joint encoders. \textcolor{black}{The robot's body, camera, and foot frames are labelled with $b$, $c$, and $f$, respectively.} ${^b}\boldsymbol{\Gamma}_{{p}}$ and ${^b}\boldsymbol{\Gamma}_{{R}}$ refer to the {translational and rotational transformations from the} body frame to {the foot frame}.}
    \label{fig:sensors}
    \vspace{-6mm}
\end{figure}

There were attempts to adopt new methodolog{ies} or framework{s}, such as Invariant \textcolor{black}{Extended Kalman Filter} (IEKF) \cite{hartley2020contact, 11teng2021legged}, analysis of legged robot’s dynamics \cite{10fourmy2021contact}, optimization on a smooth manifold \cite{14kim2021legged}, and preintegration factors \cite{8hartley2018legged, 9hartley2018hybrid, 10fourmy2021contact}.

The main advantage of utilizing IEKF is faster convergence and better performance under a wrong initial state \cite{hartley2020contact, 11teng2021legged}. Unfortunately, if the bias is augmented in the state of IEKF, it becomes ``imperfect IEKF'' \textcolor{black}{named in \cite{hartley2020contact, barrau2015non} although this imperfect IEKF outperforms the standard EKF.} 

A state estimation study has been conducted that considers not only kinematics but also dynamics more efficiently \cite{10fourmy2021contact}{, where t}he authors preintegrate the contact force. However, it is assumed that accurate contact detection is achieved, {even if} it is considered challenging.

In \cite{14kim2021legged}, the robot's state defined on a smooth manifold was optimized by using {the} Gauss-Newton method. If a slip occurs (slip can be identified through the speed of the end effector), the leg kinematic factor is ignored. 

In addition, similar to the IMU {preintegration} presented in \cite{forster2016manifold}, there were attempts to construct a more robust system by introducing the concept of the preintegrated factor \cite{8hartley2018legged, 9hartley2018hybrid, 10fourmy2021contact}. The authors of \cite{8hartley2018legged} proposed preintegrated forward kinematic factor and contact factor{, which are} adopted in this study. \cite{9hartley2018hybrid} {extended} \cite{8hartley2018legged} by adequately considering the detachment of the foot where the contact was made.

\textcolor{black}{In \cite{1camurri2020pronto}, they developed the loosely coupled state estimator utilizing not only proprioceptive sensors but also a camera and a LiDAR. The robustness and versatility of their approach were demonstrated by using several legged robots for more than two hours of total runtime.}    

Despite these {remarkable} studies, the limitations of proprioceptive sensor-based or {loosely coupled} methods are evident under extreme environment{s}. For instance, the system relying on proprioceptive sensors may diverge on a slippery surface. Similarly, the system that {uses} loosely coupled camera information easily {degraded} when significant lighting changes or repetitive patterns appear. Therefore, to develop a more robust and reliable system, there have been many recent attempts for legged robots to couple exteroceptive sensors {tightly}.

One remarkable study used factor graphs to couple visual odometry and leg odometry {tightly} \cite{13wisth2019robust}. They proposed a novel framework called {VILENS}. VILENS {did} not diverge, even if {vision degeneracy} happened. VILENS was validated on several datasets, including an environment {in which illumination was changed}. In their following study \cite{2wisth2020preintegrated}, \textcolor{black}{the body velocity can be calculated based on the non-slip assumption. And,} they empirically found that the slip effect can be modeled as a slowly time-varying bias \textcolor{black}{of the body velocity.} \textcolor{black}{The body velocities corrected by the bias are preintegrated. The preintegrated measurements constrain the two neighboring poses.} Then, \textcolor{black}{this velocity bias} was added to the state, resulting in a more {robust} system. This showed significant improvement compared to their previous approach \cite{13wisth2019robust}. These studies are incorporated and detailed in \cite{15wisth2021vilens}. \textcolor{black}{The main contribution of \cite{15wisth2021vilens} is that a camera, an IMU, joint encoders, and a LiDAR are tightly fused to achieve more robust operation when the individual sensors would otherwise degraded.}

\textcolor{black}{Our previous work, WALK-VIO \cite{lim2021walk}, tightly fuses an inertial sensor, a camera sensor, and joint encoders. The issue that the generated body motion varies with the different controllers was a motivation for developing WALK-VIO. In WALK-VIO, the walking-motion-adaptive leg kinematic constraints that change with the body motion are employed, improving the state estimator's performance.}

%%%%%%%%%%%%%%%%%%%%%%%%%%%%%%%%%%%%%% 위치 수정 됨 %%%%%%%%%%%%%%%%%%%%%%%%%%

Nevertheless, there are {still} many issues with legged robot state estimation in slippery or uneven terrain. For instance, many researchers assumed that the end effector {location} is not changed in contact {with the ground}. This approach is not appropriate if the surface is slippery. Furthermore, the {non-slip} assumption requires the contact state detection, which needs additional {sensors and} considerations. 

The contact detection methods can be classified as to whether the contact sensor is used or not. In contact sensor-based methods, %the sensor can be frequently damaged due to the force applied to the sensor, resulting in {degradation of} the whole system \cite{5hwangbo2016probabilistic}.
several limitations exist \cite{5hwangbo2016probabilistic}. First, due to the aggressive motions of legged robots and {the} impacts at the feet, the sensors would be damaged over repeated use. Second, a heavy protector for the sensor is required. Thus, to utilize the leg kinematics constraint, {detecting} the contact without using the contact sensor is preferred.

The ground reaction force (GRF) analysis is typically used for contact estimation without contact sensors. In \cite{12camurri2017probabilistic}, through a detailed analysis of GRF parameters, invalid leg odometry was discarded. Moreover, \cite{5hwangbo2016probabilistic} proposed a probabilistic way to detect contact, which {was} extended to the \textcolor{black}{Hidden Markov Model (HMM)} based probabilistic slip estimator\cite{4jenelten2019dynamic}. This approach has been demonstrated by operating ANYmal \cite{hutter2016anymal} on ice.

Even though {several studies} not using contact sensors have been {published}, an additional computation is required for any contact detection. Furthermore, the possibility of mis-detecting the contact cannot be ignored in a harsh environment, such as muddy {or} slippery surfaces. Even though previous studies assumed the slip effect {could} be modeled as Gaussian noise (or bias), this approach might be invalid under severe slip. As long as we adopt the above assumptions, the accurate contact detector is essential.

%All of the above methods share limitations. A {non-slip} assumption {is utilized} when the leg kinematics constraint is established. 

However, in this research, those assumptions are not adopted when establishing leg kinematics. Thus, the {proposed} algorithm can be utilized {even} in harsh environments, thereby broadening its application. This letter proposes a novel state estimator, STEP {(STate Estimator using Preintegrated foot velocity factor)}, {that does not rely} on an accurate contact detector and {does not assume} that the foot's position is fixed in contact. This letter makes the following contributions:

\begin{itemize}
    \item We present a novel preintegrated foot velocity factor {that} can be exploited regardless of contact state. This factor can constrain the foot pose between the consecutive image frames, which results in the improvement in the optimization of the overall cost function.
    \item The end effector velocity is estimated from leg kinematics. Note that we do not use the {non-slip} assumption. Thus, it is independent of ground characteristics, such as the friction coefficient. 
    %\item To calculate the relative foot pose, we preintegrate the foot velocity expressed in the foot frame.
    \item The performance of STEP was evaluated in harsh simulation environments and with real experimental datasets.
\end{itemize}
\vspace{-2mm}

%%%%%%%%%%%%%%%%% 나중에 분량 생기면 살리기 %%%%%%%%%%%%%%%%
%The remainder of this letter is organized as follows. Section~\ref{sec:related work} introduces related works that help understand the primary purpose and target of this letter. Section~\ref{sec:preliminaries} provides the required preliminaries including the mathematical prerequisites. We formulate the factor graph in Section~\ref{sec:problem statement}. Section~\ref{sec:foot factor} explains a novel preintegrated measurement based on forward leg kinematics. The preintegrated foot velocity factor is developed and addressed in Section~\ref{sec:foot factor}. Simulation and experimental results of the
%proposed methods on a quadruped robot (Fig. 1) are presented
%in Section~\ref{sec:experimental results}. Finally, Section~\ref{sec:conclusion} concludes and summarizes the letter in short.

%\IEEEpubidadjcol

%%%%%%%%%%%%%%%%% 3. PRELIMINARIES %%%%%%%%%%%%%%%%%
\section{Preliminaries}\label{sec:preliminaries}

In this section, preliminar{ies} of Lie groups and associated Lie Algebra are briefly presented for the following sections. More information {on} Lie group and Lie algebra can be found in \cite{forster2016manifold, LGchirikjian2009stochastic, LGabsil2009optimization, LGhall2015lie}. Especially, \cite{forster2016manifold} is recommended {for better understanding of this letter.}\vspace{-2mm}

\subsection{Useful Properties of Matrix Lie Group}\label{sec:Some Useful Properties for Matrix Lie group}

In the section, we consider the matrix Lie group $\mathcal{G}$ closed under matrix multiplication \cite{LGhall2015lie}. {Specifically}, we are interested in the rotation matrix, an element of Lie group, especially \textit{special orthogonal group}. The rotation matrix in 3D space is formally defined as follows: \vspace{-1mm}
\begin{equation}
\label{so3}
    \mathrm{SO}(3)=\left\{\mathbf{R} \in \mathrm{GL}_{3}(\mathbb{R}) \mid \mathbf{R}^{\top} \mathbf{R}=\mathbf{I}_{3\times3}, \operatorname{det} \mathbf{R}=1\right\},
    \vspace{-1mm}
\end{equation}
\noindent
where $\mathbf{R}$ is the rotation matrix, an element of $\mathrm{GL}_3(\mathbb{R})$, \textit{general linear group} of degree 3 and $\mathbf{I}_{3\times3}$ is the $3\times3$ identity matrix.

{The a}ssociated Lie algebra is denoted by $\mathfrak{g}$. The linear \textit{hat} operator, $(\cdot)^{\wedge}: \mathbb{R}^m \rightarrow \mathfrak{g}$, maps \textcolor{black}{a vector} to the Lie algebra. The tangent space to the smooth manifold, $\mathrm{SO}(3)$, (at the identity of $\mathcal{G}$) is denoted as $\mathfrak{so}(3)$, which is {the} associated Lie algebra and coincides with the space of 3 $\times$ 3 skew symmetric matrices. As described in \cite{forster2016manifold}, by adopting {the} \textit{hat} operator, Lie algebra can be \textit{vectorized} for convenience as follows: \vspace{-1mm}
\begin{equation}%\footnotesize
\boldsymbol\phi^{\wedge}=\left[\begin{array}{c}
\phi_{1} \\
\phi_{2} \\
\phi_{3}
\end{array}\right]^{\wedge}=\left[\begin{array}{ccc}
0 & -\phi_{3} & \phi_{2} \\
\phi_{3} & 0 & -\phi_{1} \\
-\phi_{2} & \phi_{1} & 0
\end{array}\right] \in \mathfrak{so}(3).
\end{equation}\normalsize

A useful property of $(\cdot)^\wedge$, \textit{anticommutative property} is introduced:     \vspace{-3mm}
\begin{equation}
    \mathbf{a}^{\wedge} \mathbf{b}=-\mathbf{b}^{\wedge} \mathbf{a} \quad \forall \mathbf{a}, \mathbf{b} \in \mathbb{R}^{3}.
    \label{eq:anticomm}
    \vspace{-2mm}
\end{equation}\normalsize

For the corresponding inverse map, \textit{vee} operator is introduced as $(\cdot)^{\vee}: \mathfrak{g} \rightarrow \mathbb{R}^m$. {The \textit{hat} operator and the \textit{vee} operator have the following relationship}$: \mathbf{R} = {\mathbf{a}}^\wedge$ then $\mathbf{R}^\vee = \mathbf{a}$. More information can be found in \cite{forster2016manifold}.

The exponential map, $\operatorname{exp}: \mathfrak{so}(3) \rightarrow \mathrm{SO}(3)$, is defined as follows:
\vspace{-1mm}\small
\begin{equation}
\exp \left(\boldsymbol\phi^{\wedge}\right)=\mathbf{I}_{3\times3}+\frac{\sin (\|\boldsymbol\phi\|)}{\|\boldsymbol\phi\|} \boldsymbol\phi^{\wedge}+\frac{1-\cos (\|\boldsymbol\phi\|)}{\|\boldsymbol\phi\|^{2}}\left(\boldsymbol\phi^{\wedge}\right)^{2}.
\end{equation}\normalsize

It is often approximated that $\exp \left(\boldsymbol\phi^{\wedge}\right) \approx \mathbf{I}_{3\times3}+\boldsymbol\phi^{\wedge}$, where $\boldsymbol\phi\approx0$. 

Likewise, for any $\|\boldsymbol\phi\|<\pi$, {the} logarithm map, $\operatorname{log}: \mathrm{SO}(3) \rightarrow \mathfrak{so}(3)$, which associates a Lie group element $\mathbf{R}\neq \mathbf{I}_{3\times3}$ in $ \mathrm{SO}(3)$ to a Lie algebra element is defined as follows:
\begin{equation}\small
\log (\mathbf{R})=\frac{\varphi \cdot\left(\mathbf{R}-\mathbf{R}^{\top}\right)}{2 \sin (\varphi)} \text { with } \varphi=\cos ^{-1}\left(\frac{\operatorname{tr}(\mathbf{R})-1}{2}\right).
\end{equation}\normalsize

{Note that it {represents} the rotation by using the rotation axis and {the} rotation angle{:} $\operatorname{log}(\mathbf{R})^{\vee} = \mathbf{a}\phi$, where $\mathbf{a}$ and $\phi$ are the rotation axis and the rotation angle of $\mathbf{R}$, respectively. If $\mathbf{R} = \mathbf{I}_{3\times3}$, then the rotation angle, $\phi = 0$ and the rotation axis, $\mathbf{a}$, can be chosen arbitrarily.}

Similar to the {vectorization of} Lie algebra above, the exponential and logarithm map is \textit{vectorized} as below \cite{forster2016manifold}: \vspace{-1mm}
\begin{equation}%\small
\begin{aligned}
&\operatorname{Exp}: \mathbb{R}^{3} \rightarrow \mathrm{SO}(3)\quad ; \boldsymbol\phi \mapsto \exp \left(\boldsymbol\phi^{\wedge}\right) \\
&\operatorname{Log} : \mathrm{SO}(3) \rightarrow \mathbb{R}^{3} \quad ; \mathbf{R} \mapsto \log (\mathbf{R})^{\vee}.
\end{aligned}
\vspace{-1mm}
\end{equation}\normalsize

Later, for small $\delta\boldsymbol\phi$, the following first-order approximation will be used:
\vspace{-2mm}\begin{equation}\vspace{-1mm}%\small
\operatorname{Exp}(\boldsymbol\phi+\delta\boldsymbol\phi) \approx \operatorname{Exp}(\boldsymbol\phi) \operatorname{Exp}\left(\mathbf{J}_{r}(\boldsymbol\phi) \delta \boldsymbol\phi\right),
\label{eq:ExpJac}
\end{equation}%\normalsize
\begin{equation}%\small
\vspace{-1mm}
\operatorname{Log} (\operatorname{Exp}(\boldsymbol\phi) \operatorname{Exp}(\delta \boldsymbol\phi)) \approx \boldsymbol\phi+\mathbf{J}_{r}^{-1}(\boldsymbol\phi) \delta \boldsymbol\phi,
\end{equation}
\begin{equation}%\small
\operatorname{Exp}(\delta \boldsymbol\phi) \approx \mathbf{I}+(\delta \boldsymbol\phi)^{\wedge},
\label{eq:Expassume}\vspace{-1mm}
\end{equation}%\normalsize
\noindent
where $\mathbf{J}_{r}(\boldsymbol\phi)$ is {right Jacobian}. The derivation of $\mathbf{J}_{r}(\boldsymbol\phi)$ can be found in \cite{LGchirikjian2009stochastic, forster2016manifold} {as follows:}\vspace{-1.0mm}
\begin{equation}\small
\mathbf{J}_{r}(\boldsymbol\phi)=\mathbf{I}-\frac{1-\cos (\|\boldsymbol\phi\|)}{\|\boldsymbol\phi\|^{2}} \boldsymbol\phi^{\wedge}+\frac{\|\boldsymbol\phi\|-\sin (\|\boldsymbol\phi\|)}{\left\|\boldsymbol\phi^{3}\right\|}\left(\boldsymbol\phi^{\wedge}\right)^{2}.
\label{eq:rightJac}\normalsize
\end{equation}

Lastly, another property of {the} exponential map of $\mathrm{SO}$(3) is introduced:\vspace{-1mm}
\begin{equation}\vspace{-1mm}%\small
\mathbf{R}\operatorname{Exp}(\boldsymbol\phi) \mathbf{R}^{\top}=\operatorname{exp} \left(\mathbf{R} \boldsymbol\phi^{\wedge} \mathbf{R}^{\top}\right)=\operatorname{Exp}(\mathbf{R} \boldsymbol\phi),
\end{equation}
\begin{equation}
\operatorname{Exp}(\boldsymbol\phi) \mathbf{R}=\mathbf{R} \operatorname{Exp}\left(\mathbf{R}^{\top} \boldsymbol\phi\right).\label{eq:commExp}
%\normalsize
\end{equation}

%Lastly, a useful representation is introduced, \textit{adjoint}, $(\text{Ad}(\mathbf{T})\mathbf{\boldsymbol{\xi}})^{\wedge}=\mathbf{T}\boldsymbol{\xi}^{\wedge}\mathbf{T}^{-1}$ for $\mathbf{T} \in \mathcal{G}$ and $\boldsymbol\xi \in \mathbb{R}^m$.

\subsection{Uncertainty Description on a Smooth Manifold}\label{sec:uncertainty description}

One advantage of representing the rotation as the Lie group and {the} associated Lie algebra is that uncertainty can be described without losing the Gaussian property \cite{forster2016manifold}. The uncertainty in $\mathrm{SO}(3)$ is modeled by defining a noise distribution in the tangent space, its Lie algebra $\mathfrak{so}(3)$, and then mapping it to $\mathrm{SO}(3)$ through the exponential map \cite{8hartley2018legged, forster2016manifold}, which will be explained in the following section. The perturbed rotation matrix can be written as follows:\vspace{-1mm}
\begin{equation}%\small
\tilde{\mathbf{R}}=\mathbf{R} \operatorname{Exp}\left(\delta \boldsymbol\phi\right), \quad \delta\boldsymbol\phi \sim \mathcal{N}(0, \boldsymbol\Omega),
\label{eq:uncerRot}
\vspace{-1mm}
\end{equation}
\noindent
where $\delta\boldsymbol\phi$ is a normally distributed small perturbation with zero mean and covariance $\boldsymbol\Omega$, {and} ${\mathbf{R}}$ {is} the noise-free rotation. The detailed derivation of the distribution of $\mathbf{R}$ {is indicated in} \cite{barfoot2014associating, forster2016manifold}. {We adopt the result of the negative log-likelihood {$\mathcal{L}$} of a rotation $\mathbf{R}$ given a noisy measurement $\tilde{\mathbf{R}}$: 
\begin{equation}%\small
%\vspace{-1mm}
\mathcal{L}({\mathbf{R}}) \propto \frac{1}{2}\left\|\operatorname{Log} \left({\mathbf{R}}^{-1} \tilde{\mathbf{R}}\right)\right\|_{\boldsymbol\Sigma}^{2} = \frac{1}{2}\left\|\operatorname{Log} \left(\tilde{\mathbf{R}}^{-1} {\mathbf{R}}\right)\right\|_{\boldsymbol\Sigma}^{2}.
\end{equation}}\normalsize

The uncertainty of translation can be characterized by exploiting the additive Gaussian noise assumptions \cite{forster2016manifold}.\vspace{-2mm}

\subsection{Optimization on a Smooth Manifold}\label{sec:Optimization on Smooth Manifold}
%The optimization method on a smooth manifold, $\mathcal{G}$, {often} attracts research interest. Unfortunately, 
We could not directly apply vector calculus to the body orientation involved in the state that evolves on the $\mathrm{SO}(3)$ manifold. Thus, we adopt the approach suggested in \cite{forster2016manifold, 14kim2021legged}, called the \textit{lift-solve-retract} scheme. Furthermore, for the required retraction for $\mathrm{SO}$(3) and lifting for $\mathfrak{so}$(3), $\operatorname{Exp}(\cdot)$ and $\operatorname{Log}(\cdot)$ map{s are adopted}, which are introduced in Section \ref{sec:Some Useful Properties for Matrix Lie group}.\vspace{-2mm}

\section{Factor Graph Formulation}\label{sec:problem statement}
In this section, we explain the factor graph formulation of STEP. The factor graph is based on that of {the} VINS-Fusion~\cite{qin2019general} framework, which is a tightly coupled, sliding{-}window nonlinear optimization-based VIO algorithm. To construct a state estimator for legged robots, we added a novel preintegrated foot velocity factor to the factor graph. In addition, we modified VINS-Fusion to optimize the factor graph on-manifold.
As shown in Fig.~\ref{fig:sensors}, a sensor configuration consisting of a stereo camera, {an} IMU sensor, and joint encoders {for} each leg {was} used, and the contact sensor {was} not used. The body frame {was} located on {the} IMU. \vspace{-2mm}
\subsection{State Definition}\label{sec:state}
The state vector used in this research {is as follows:}%\vspace{-2mm}
\begin{equation}
\begin{split}
&\mathcal{X} = [\mathbf{x}_{0}, \mathbf{x}_{1}, \cdots ,\mathbf{x}_{i-1}, \lambda_{0}, \lambda_{1}, \cdots, \lambda_{k}],\\
&\mathbf{x}_i = [\mathbf{p}^{w}_{b_i}, \mathbf{R}^{w}_{b_i}, \mathbf{v}^{w}_{b_i}, \mathbf{\Psi}^w_{l,i}, \mathbf{s}^w_{l,i}, \mathbf{b}^{a}_{i}, \mathbf{b}^{g}_{i}],
\label{eq:state}
\end{split}
\vspace{-3mm}
\end{equation}
\noindent where $\mathbf{x}_{i}$ represents the robot state when the $i$-th keyframe is input. It contains the body position, $\mathbf{p}^{w}_{b_i}\in\mathbb{R}^3${;} orientation $\mathbf{R}^{w}_{b_i}\in\mathrm{SO}(3)${;} velocity, $\mathbf{v}^{w}_{b_i}\in\mathbb{R}^3${;} orientation of {the} $l$-th end effector, $\mathbf{\Psi}^w_{l,i}${;} position of {the} $l$-th end effector, $\mathbf{s}^w_{l,i}${;} {the} IMU accelerometer and gyroscope biases, $\mathbf{b}^{a}_{i}\in\mathbb{R}^3$ and $\mathbf{b}^{g}_{i}\in\mathbb{R}^3${;} and $\lambda_{k}$ indicates the $k$-th inverse depth of the viusual feature in the first observed camera frame. {Note that the superscript $w$ denotes the parameters are estimated in the world frame.} {For readability throughout this letter}, {however,} the world frame superscripts {will be} dropped. For example, $\mathbf{p}^{w}_{b_i}$, $\mathbf{v}^{w}_{b_i}$, and $\mathbf{R}^{w}_{b_i}$ will be abbreviated as ${\mathbf{p}_i}$, ${\mathbf{v}_i}$, and $\mathbf{R}_i$, respectively. If it is not represented in the world frame, then a reference frame is denoted {by a} left side superscript, such as the body frame ${^b}(\cdot)$, the camera frame ${^c}(\cdot)$, and the foot frame ${^f}(\cdot)$. Lastly, $\tilde{(\cdot)}$ denotes the noisy measurement.

\begin{figure}[t]
    \centering
    \includegraphics[width=0.9\linewidth]{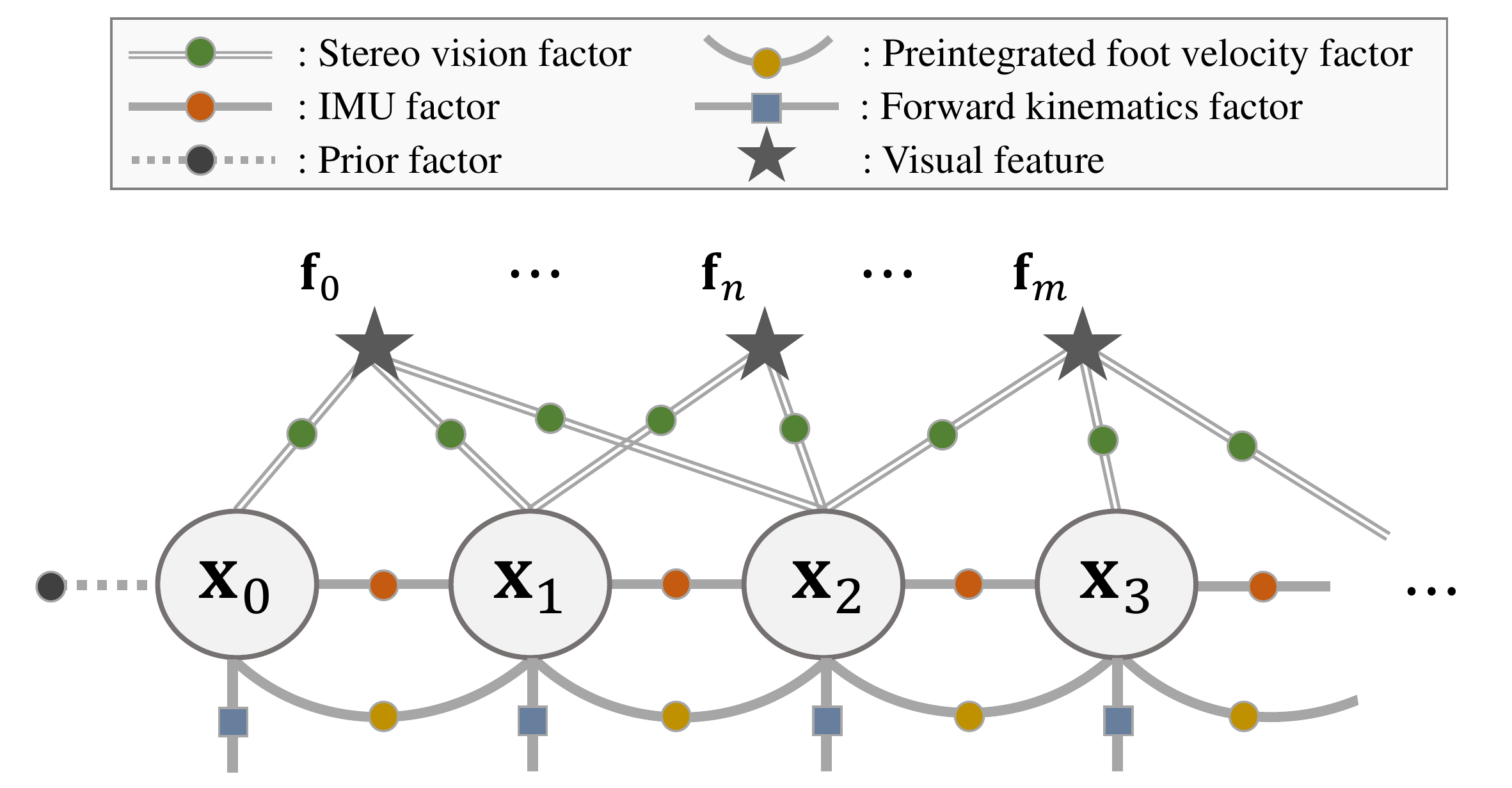}
    \caption{The visualization of the factor graph structure. The factor graph consists of measurement factors: (1) prior factor, (2) IMU factor, (3) visual factor, (4) forward kinematics factor, and (5) preintegrated foot velocity factor. Unlike the forward kinematics factor, the preintegrated foot velocity {factor} can relate the foot pose between the consecutive frames.}%{, just} as {the} IMU factor does.}
    \label{fig:factor graph}
    \vspace{-3mm}
\end{figure}

\subsection{Measurements and Factor Graph}\label{sec:factor graph}
The sensors used in this letter are a stereo camera, IMU, and leg joint encoders. All measurements from each sensor up to {the} $k$-th keyframe, $\mathcal{Z}_{k}$, can be represented as follows:\vspace{-1mm}
\begin{equation}
\tiny
    \mathcal{Z}_{k} = \displaystyle\bigcup_{\forall{(i,j)}\in\mathcal{T}_{k}}\{\mathcal{I}_{ij}, \mathcal{C}_{i}, {\mathcal{K}}^l_{i}, {\mathcal{V}}^l_{ij}\}, l=\left\{1,\ldots N\right\}, \nonumber\vspace{-1.5mm}
\label{eq:measurement}
\end{equation}
\noindent where $\mathcal{T}_{k}$ is the set of timestamps of the keyframe in the $k$-th sliding window. We assume the IMU and joint encoder{s} are synchronized with the camera. {Furthermore,} $\mathcal{I}_{ij}\in\mathbb{R}^6$ refers to the {preintegrated} IMU measurement between timestamps $i$ and $j${;} $\mathcal{C}_{i}$ is the keyframe obtained at time $i${;} $\mathcal{K}^{l}_{i}\in\mathbb{R}^{n_\mathrm{joints}}$ is the forward kinematic measurement of {the} $l$-th leg at time $i$, {where $n_\mathrm{joints}$ denotes the number of joints}. $\mathcal{V}^{l}_{ij}\in\mathbb{R}$ denotes the foot velocity measurement of the $l$-th leg between $i$ and $j$, which {is} obtained from leg kinematics.

The IMU and joint encoder measurements are input at higher frequencies. So, {the IMU measurements are preintegrated,} as proposed in\cite{forster2016manifold}. {For similar reasons, the joint measurements are preintegrated} and used for tracking the pose of the $l$-th end effector between two frames. A detailed description of the preintegration of foot velocity will be given in Section \ref{sec:foot factor}. \vspace{-4mm}

\subsection{Cost Function}\label{sec:cost function}
If all sensor measurements are conditionally independent to each other, the {maximum} posterior state $\mathcal{X}_k$, given {the} measurement $\mathcal{Z}_k$, can be expressed as: \vspace{-1mm}
\begin{equation}
\mathcal{X}^*_k = \operatorname*{arg\,max}_{\mathcal{X}_k} p\big{(}\mathcal{X}_k\big{|}\mathcal{Z}_{k}\big{)} \propto p\big{(}\mathcal{X}_0\big{)}p\big{(}\mathcal{Z}_k\big{|}\mathcal{X}_{k}\big{)},
\label{eq:MAP}
\vspace{-1mm}
\end{equation}
\noindent where \footnotesize$p\big{(}\mathcal{Z}_k\big{|}\mathcal{X}_{k}\big{)}=\prod_{(i,j)\in\mathcal{T}_k}p\big{(}\mathcal{I}_{ij}\big{|}\mathcal{X}_{j}\big{)}p\big{(}\mathcal{C}_{i}\big{|}\mathcal{X}_{i}\big{)}p\big{(}\mathcal{K}_{i}\big{|}\mathcal{X}_{i}\big{)p\big{(}\mathcal{V}_{ij}\big{|}\mathcal{X}_{j}\big{)}}.$\normalsize

{(\ref{eq:MAP}) can be transformed into an {equivalent nonlinear} least-square problem. Therefore, the final {objective} we used for obtaining a maximum posteriori estimate of $\mathcal{X}_k$ can be formulated as follows:}\par\vspace{-4mm}
\small
\begin{equation}
\begin{gathered}
\min_{\mathcal{X}_k} \Bigg\{
\parallel \textcolor{black}{\mathbf{r}_p - \mathbf{H}_p\mathcal{X}_k}\parallel^2 
+ \sum_{(i,j) \in \mathcal{T}_k}\parallel\textcolor{black}{{\mathbf{r}_{\mathcal{I}}(\mathcal{I}_{ij},\mathcal{X}_k)}}\parallel_{\mathbf{\Omega}_{\mathcal{I}_{ij}}}^2 +\\
 \sum_{i \in \mathcal{T}_k}\sum_{n \in \mathcal{F}}\rho(\parallel\textcolor{black}{{\mathbf{r}_{\mathcal{C}}(\mathcal{C}_{i,\mathbf{f}_n},\mathcal{X}_k)}}\parallel_{\mathbf{\Omega}_{\mathcal{C}_{i,n}}}^2) + \sum_{i \in \mathcal{T}_k}\sum_{l=1}^{N}\parallel\textcolor{black}{{\mathbf{r}_{\mathcal{K}}(\mathcal{K}^{l}_{i},\mathcal{X}_k)}}\parallel_{\mathbf{\Omega}_{\mathcal{K}_{i,l}}}^2\\
+ \sum_{(i,j) \in \mathcal{T}_k}\sum_{l=1}^{N}\parallel\textcolor{black}{{\mathbf{r}_{\mathcal{V}}(\mathcal{V}^{l}_{ij},\mathcal{X}_k)}}\parallel_{\mathbf{\Omega}_{\mathcal{V}_{ij,l}}}^2 \Bigg\},
\label{eq:total residual}
\end{gathered}
\end{equation} \normalsize
\noindent where $\mathcal{F}$ is the set of visual feature indices. Each sensor noise covariance is expressed as $\mathbf{\Omega}_{\mathcal{I}_{ij}}, \mathbf{\Omega}_{\mathcal{C}_{i,n}}$, $\mathbf{\Omega}_{\mathcal{K}_{i,l}}$, and $\mathbf{\Omega}_{\mathcal{V}_{ij,l}}$. In addition, $\parallel\cdot\parallel$ refers to the Mahalanobis norm, and $\rho(\cdot)$ {refers to} the Huber norm~\cite{huber1992robust}. The cost function is defined as the sum of each measurement factor: (1) prior factor $\mathbf{r}_0$, (2) IMU factor $\mathbf{r}_{\mathcal{I}}$, (3) visual factor $\mathbf{r}_{\mathcal{C}}$, (4) forward kinematics factor $\mathbf{r}_{\mathcal{K}}$, and (5) preintegrated foot velocity factor $\mathbf{r}_{\mathcal{V}}$. %\textcolor{black}{The prior factor and visual factor are adopted from \cite{qin2019general}. The IMU factor {is from} \cite{forster2016manifold}, but we do not consider the bias update between consecutive keyframes {due to high computational complexity compared to its gain.}} 
\textcolor{black}{For solving the nonlinear least square problem, Levenberg-Marquardt algorithm\cite{more1978levenberg} is used.} The visualization of the corresponding factor graph {is shown} in Fig. ~\ref{fig:factor graph}.
\vspace{-2mm}
\textcolor{black}{\subsection{VIO Factors}}\label{sec:factors} \vspace{-1mm}
\textcolor{black}{In this section, each factor forming the cost function is described. The prior factor and visual factor are adopted from \cite{qin2019general}. The IMU factor {is from} \cite{forster2016manifold}. The leg kinematic-related factors such as the forward kinematics factor and preintegrated foot velocity factor will be explained in Section~\ref{sec:foot factor}.}

\textit{\textcolor{black}{1) Prior factor}}:
\textcolor{black}{Due to the computational cost, a sliding window-based method is adopted. Therefore, a prior factor is used to serve as an anchor whenever marginalization occurs. The prior factor $\mathbf{r}_0$ is defined as the error between the prior state $\mathbf{x}_0$ and the estimated prior state $\mathbf{\hat{x}}_0$.}

\textcolor{black}{\textit{2) IMU factor}: Since the IMU usually offers data with a higher frequency than the camera, the measurements between the frames are preintegrated to compute the changes in pose $\Delta\mathbf{\tilde{p}}$, velocity $\Delta\mathbf{\tilde{v}}$, and orientation $\Delta\mathbf{\tilde{R}}$ of the robot. And these changes can constrain the two neighboring nodes of the graph by defining the IMU factor $\mathbf{r}_{\mathcal{I}}$ given IMU measurements $\mathcal{I}_{ij}$ as follows: \vspace{-1.5mm}
\begin{equation}
\footnotesize
\mathbf{r}_{\mathcal{I}}(\mathcal{I}_{ij},\mathcal{X}_k)=
% \begin{pmatrix}
% \mathbf{r}_{\Delta\mathbf{p}_{ij}}\\
% \mathbf{r}_{\Delta\mathbf{v}_{ij}}\\
% \mathbf{r}_{\Delta\mathbf{R}_{ij}}\\
% \mathbf{r}_{\Delta\mathbf{b}^{a}_{ij}}\\
% \mathbf{r}_{\Delta\mathbf{b}^{g}_{ij}}
% \end{pmatrix}
% =
\begin{pmatrix}
\mathbf{R}^{\top}_{i}(\mathbf{p}_{j}-\mathbf{p}_{i}-\mathbf{v}\Delta t_j-\frac{1}{2}\mathbf{g}\Delta t^2_j)-\Delta\tilde{\mathbf{p}}_j\\
\mathbf{R}^{\top}_{i}(\mathbf{v}_{j}-\mathbf{v}_{i}-\mathbf{g}\Delta t_j)-\Delta\tilde{\mathbf{v}}_j\\
\operatorname{Log}\big{(}\Delta{\tilde{\mathbf{R}}_j}\mathbf{R}^{\top}_{i}\mathbf{R}_{j}\big{)}\\
\mathbf{b}^{a}_{j} - \mathbf{b}^{a}_{i}\\
\mathbf{b}^{g}_{j} - \mathbf{b}^{g}_{i}
\end{pmatrix}.
\vspace{-1mm}
\end{equation}}
\normalsize The IMU factor is \textcolor{black}{adopted} from \cite{forster2016manifold}, but we do not consider the bias update \textcolor{black}{between consecutive keyframes} due to computational complexity.

%Unlike \cite{forster2016manifold}, the bias update is not considered due to high computational complexity.}

\textcolor{black}{\textit{3) Visual factor}: For the visual factor, the traditional reprojection error is utilized. The visual factor $\mathbf{r}_{\mathcal{C}}$ when the 3D feature $\mathbf{f}_n$ observed in the $i$-th frame is defined as: \vspace{-1mm}
\begin{equation}
\mathbf{r}_{\mathcal{C}}(\mathcal{C}_{i,\mathbf{f}_n},\mathcal{X}_k)=
\textbf{\emph{f}}_{i,n} - \pi\big{(}\mathbf{R}_i,\mathbf{p}_i,\mathbf{f}_n \big{)}, 
\vspace{-1mm}
\end{equation}
\noindent where $\pi(\cdot)$ is the projection function which projects the 3D feature to the image and $\textbf{\emph{f}}_{i,n} $ is the $n$-th feature observed in the $i$-th frame.}
\vspace{-2mm}
\section{Leg Factors}\label{sec:foot factor} \vspace{-1mm}
%\textcolor{black}{왜냐하면 (world frame에서 기술된) 발 끝의 속도와 자세 정보가 severe slip에 의해 제대로 추정되지 않는다면, 이와 tightly coupled된 ego-motion state도 deteriorate될 수 있기 때문이다. Moreover, by suggesting a novel preintegrated factor, 더욱 효율적인 계산이 가능하기에다. 단순히 noise 또는 bias로 모델링하는 것이 아니라, IMU와 joint encoder의 measurements를 사용하여 발 끝의 속도와 자세를 추정하는 것이므로, the slip effect can be modeled more precisely.} 
This section presents the novel preintegrated foot velocity factor, which is our key contribution, and the forward kinematics factor. As we introduced in Section~\ref{sec:introduction}, many researchers have proposed various leg factors constrained by non-slip assumption. \textcolor{black}{Additionally, most of them considered the effect of sensor noises or slippage as a Gaussian noise or bias. However, those techniques might fail to estimate the end effector pose in a severe slippage condition, which leads to deteriorating the estimated body pose tightly coupled to the end effector pose.} In addition, they can constrain the pose only when the contact state is maintained for a certain period, and it would be vulnerable to the slip occurrences. Therefore, we propose a preintegrated foot velocity factor that does not depend on the contact state and {that} can be used at all times. For this, the changes in foot pose between consecutive frames are calculated by preintegrating the end effector velocities. \vspace{-2mm}  

%these approaches require estimating the contact state or installation of the contact sensor.   

\subsection{Measurement Model}\label{sec:state}
In this section, foot linear and angular velocit{ies} are {derived} in the foot frame, which is necessary for later preintegration of foot linear and angular velocity. {To this end}, forward kinematics of a legged robot and transformation between frames are exploited.

The joint encoder measurement vector $\tilde{\boldsymbol\alpha}(t)$ can be expressed as follows:\vspace{-2mm}
\begin{equation}%\small
\tilde{\boldsymbol{\alpha}}(t)= \boldsymbol{\alpha}(t) + \mathbf{n}^{\alpha}(t)\in\mathbb{R}^M, \vspace{-2mm}
\end{equation} 
\noindent where $\boldsymbol\alpha(t)$ is the true joint angle{;} $\mathbf{n}^{\alpha}(t)\sim\mathcal{N}(0,\boldsymbol{\Omega}^{\alpha})$ {is} the joint measurement noise modeled as a Gaussian noise with covariance $\boldsymbol{\Omega}^{\alpha}${;} and $M$ is the number of joint encoders of each leg.

When $\tilde{\boldsymbol{\alpha}}(t)$ is given, by using the leg kinematic{s} model, {the} foot position $\mathbf{s}(t)\in\mathbb{R}^3$ and orientation $\mathbf{\Psi}(t)\in\mathrm{SO}(3)$ in the world frame at $t$ can be calculated as follows: \vspace{-1mm}
\begin{equation}%\small
\mathbf{\Psi}(t)=\mathbf{R}(t)\,{^b}\boldsymbol{\Gamma}_{{R}}(\boldsymbol{\alpha}(t))=\mathbf{R}(t)\,{^b}\boldsymbol{\Gamma}_{{R}}(\tilde{\boldsymbol{\alpha}}(t)-\mathbf{n}^{\alpha}(t))
\label{eq:foot pos}
\end{equation}
\vspace{-4mm}
\begin{equation} %\small
\begin{split}
    \mathbf{s}(t)&=\mathbf{R}(t)\,{^b}\boldsymbol{\Gamma}_{{p}}(\boldsymbol{\alpha}(t))+\mathbf{p}(t)\\
    &=\mathbf{R}(t){^b}\boldsymbol{\Gamma}_{{p}}(\tilde{\boldsymbol{\alpha}}(t)-\mathbf{n}^{\alpha}(t))+\mathbf{p}(t), 
\end{split}
\label{eq:foot position}
\end{equation}
\noindent where ${^b}\boldsymbol{\Gamma}_p(\cdot)$ and ${^b}\boldsymbol{\Gamma}_R(\cdot)$ are the end effector position and orientation calculated by forward kinematics, respectively \cite{8hartley2018legged}. Note that ${^b}\boldsymbol{\Gamma}_p(\cdot)$ and ${^b}\boldsymbol{\Gamma}_R(\cdot)$ are expressed in the body frame. %the orientation $\boldsymbol{\Gamma}_{{R}}(\cdot)$ of foot relative to body can be calculated as\cite{8hartley2018legged}:
 %%%%%%%%%% alpha dagger 썼던 식 %%%%%%%%%%
\begin{comment}
\begin{align}
\begin{split}
    &\boldsymbol{\Gamma}_{R}(\boldsymbol{\alpha}(t)) =  \prod_{k=1}^{N-1}\mathbf{A}_k(t)\operatorname{Exp}(\boldsymbol{\alpha}_k^{\dagger}(t)) \\
    %&\mathbf{fk}_{p}(\boldsymbol{\alpha})=\sum_{k=1}^{N} \mathbf{A}_{k}(\boldsymbol{\alpha}) \mathbf{t}_{k}.
\end{split}
\end{align} 
$\mathbf{A}_k(t)\in\mathrm{SO}(3)$ represents the rotation defined from the kinematic model, $N$ the number of links, $\alpha_k^\dagger(t)$ the vector representation of rotation axis and angle. For instance, $\alpha_k^\dagger(t) = [\alpha(t), 0, 0]^T$ represents the rotation axis is $x$-axis and rotation angle is $\alpha(t)$.
\end{comment}
%%%%% %%%%%%%%%% %%%%%%%%%% %%%%%%%%%% %%%%%%%%%%
From now on, we omit $t$ for {brevity}. 

To represent the foot angular velocity in the foot frame, we differentiate (\ref{eq:foot pos}) on both sides:\vspace{-1mm}
\begin{equation}
%\begin{split}
    \dot{\boldsymbol\Psi} =   \mathbf{R}{(^{b}{\mathbf{w}})}^\wedge\,{^b}\boldsymbol\Gamma_R({\boldsymbol\alpha}) + \mathbf{R}\,{^b}\boldsymbol\Gamma_R({\boldsymbol\alpha}){({^b}\boldsymbol\omega)}^\wedge,
%\end{split} 
\label{eq:psiDot}
\vspace{-1mm}
\end{equation}
\noindent where $^{b}{\mathbf{w}}$ is the body angular velocity, {and} ${^b}\boldsymbol\omega$ the foot angular velocity expressed in the body frame. Note that ${^b}\boldsymbol\omega$ can be computed because of leg kinematics.
%\begin{equation}
 %   {^b}\boldsymbol\omega = {(\boldsymbol\Gamma_R^{\top}\mathbf{J}_R({\boldsymbol\alpha})\dot{\boldsymbol\alpha})}^\vee
%\end{equation}
%\noindent where $\mathbf{J}_R({\boldsymbol\alpha})=\frac{\delta{^b}\boldsymbol\Gamma_{R}\left({\boldsymbol\alpha}\right)}{\delta{\boldsymbol{\alpha}}}$.}
%%%%%%%%%%% Previous eq %%%%%%%%%%%%%%
%where $\mathbf{J}_R(\tilde{\boldsymbol\alpha})=\frac{\delta{^b}\boldsymbol\Gamma_{R}\left(\tilde{\boldsymbol\alpha}\right)}{\delta\tilde{\boldsymbol{\alpha}}}$, $^{b}\tilde{\mathbf{w}}$ is the IMU gyro measurement at $t$, $\mathbf{n}^{{\dot{\boldsymbol{\Psi}}}}$ the single noise term that combines gyro measurement noise, joint encoder noise, and imprecise kinematic modeling.  

Alternatively, the derivative of $\boldsymbol\Psi$ can be represented as the multiplication of foot orientation and foot angular velocity:\vspace{-1mm}
\begin{equation}
    \dot{\boldsymbol\Psi} = \boldsymbol\Psi{({^f}{\boldsymbol\omega})}^{\wedge},
\label{eq:contPsi}
\vspace{-1mm}
\end{equation}
\noindent where ${^f}{\boldsymbol\omega}$ is the foot angular velocity represented in the foot frame. 

Using~(\ref{eq:psiDot}) and~(\ref{eq:contPsi}), we can write ${^f}{\boldsymbol\omega}$ as a function of ${^b}{\mathbf{w}}$, ${\boldsymbol\alpha}$, and $^b{{\boldsymbol\omega}}$. For the sake of simplicity, we omit ${\boldsymbol\alpha}$ from now on:%\vspace{-3mm}
\begin{equation}
        {^f}{\boldsymbol\omega}={({^b}\boldsymbol\Gamma_R^\top{({{^b}{\mathbf{w}}})}^{\wedge}\,{^b}\boldsymbol\Gamma_R+ {(^b{\boldsymbol\omega})}^\wedge)}^{\vee}.
        %\vspace{-1mm}
\label{eq:footAng}
\end{equation}
\indent(\ref{eq:footAng}) can be written with sensor measurements as follows: %\vspace{-2mm}
\begin{equation}
        {^f}{\tilde{\boldsymbol\omega}}={({^b}\boldsymbol\Gamma_R^\top{({{^b}{\tilde{\mathbf{w}}-\mathbf{b}^g}})}^{\wedge}\,{^b}\boldsymbol\Gamma_R+{(^b{\tilde{\boldsymbol\omega}})}^\wedge)}^{\vee} + \mathbf{n}^{\tilde{\boldsymbol\omega}},
        \label{eq:footAngwNoise}%\vspace{-1mm}
\end{equation}
\noindent where $\mathbf{n}^{\tilde{\boldsymbol\omega}}$ {is the single noise term} with covariance $\boldsymbol\Omega^{{\tilde{\boldsymbol\omega}}}$ that combines the effects of gyro measurement noise, joint encoder noise, and imprecise kinematic modeling. This strategy is inspired by \cite{3bloesch2013state} and \cite{8hartley2018legged}.  

Likewise, $\dot{\mathbf{s}}$ can be interpreted as a foot linear velocity expressed in the world frame, written as follows:\vspace{-1mm}
\begin{equation}
    \dot{\mathbf{s}}  = {\boldsymbol\nu} =\mathbf{R}{({^b}{\mathbf{w}})}^\wedge\, {^b}\boldsymbol{\Gamma}_{p}+\mathbf{R}\mathbf{J}_p{{{\dot{\boldsymbol{\alpha}}}}} + \mathbf{v},
\label{eq:foot velocity world}
\vspace{-1mm}
\end{equation}
\noindent where ${\boldsymbol\nu}$ is the foot velocity expressed in the world frame{;} $\mathbf{J}_p=\frac{\delta{^b}\boldsymbol\Gamma_{p}\left({\boldsymbol\alpha}\right)}{\delta{\boldsymbol{\alpha}}}${;} and $\mathbf{v}$ {is} the body velocity represented in the world frame.

Note that (\ref{eq:foot velocity world}) should be transformed to the foot frame and expressed with sensor measurements to  {find the} preintegrated foot measurement. We manipulate (\ref{eq:foot velocity world}) by multiplying $\boldsymbol\Psi^{\top}$ to both sides and augmenting the measurements, leading to: %\vspace{-2mm}
\begin{equation}
\begin{split}
        {^f}\tilde{\boldsymbol\nu} &= {^b}\boldsymbol\Gamma_R^{\top}{(^{b}\tilde{\mathbf{w}}-\mathbf{b}^{g})}^\wedge\, {^b}\boldsymbol{\Gamma}_{p}+{^b}\boldsymbol\Gamma_R^{\top}\mathbf{J}_p{{\tilde{\dot{\boldsymbol{\alpha}}}}} + {^b}\boldsymbol\Gamma_R^{\top}\mathbf{R}^\top\mathbf{v}+\mathbf{n}^{\tilde{\boldsymbol\nu}}\\
        &\simeq {^b}\boldsymbol\Gamma_R^{\top}{(^{b}\tilde{\mathbf{w}}-\mathbf{b}^{g})}^\wedge\, {^b}\boldsymbol{\Gamma}_{p}+{^b}\boldsymbol\Gamma_R^{\top}\mathbf{J}_p{{\tilde{\dot{\boldsymbol{\alpha}}}}}+ {^f}\tilde{\mathbf{v}} + \mathbf{n}^{\tilde{\boldsymbol\nu}},
        \label{eq:foot velocity foot} \raisetag{18pt}
\end{split}
\end{equation}
\noindent where ${^f}\tilde{\mathbf{v}}$ is the {body velocity measurement} transformed {to} the foot frame and $\mathbf{n}^{\tilde{\boldsymbol\nu}}$ is the noise term of ${^f}\tilde{\boldsymbol\nu}$ with covariance $\boldsymbol{\Omega}^{\tilde{\boldsymbol\nu}}$, which can be written as in (\ref{eq:footAngwNoise}). Note that we assume that $\mathbf{n}^{\tilde{\boldsymbol\omega}}$ and $\mathbf{n}^{\tilde{\boldsymbol\nu}}$ are Gaussian white noises. 

For foot velocity preintegration, we stress that $\mathbf{R}^{\top}\mathbf{v}$ in~(\ref{eq:foot velocity foot}) can be approximated by the body velocity measurement obtained {from a stereo vision with optical flow analysis} as follows\textcolor{black}{~\cite{ci2016robust}}:  %\textcolor{black}{2D features observed by both the current and previous frames of the stereo camera are triangulated. Then, by differentiating the 3D point feature positions at each frame, the body velocity can be calculated as follows:} 
%\vspace{-1.5mm}
\begin{equation}
    ^{b}\tilde{\mathbf{v}}=\mathbf{R}_c^b~{^c}\tilde{\mathbf{v}},
%= -\mathbf{R}_c^b~(\frac{d{^c}\bold{{f}}}{dt} + \mathbf{R}_b^c({{^b}{\tilde{\mathbf{w}}-\mathbf{b}^g})}^{\wedge}\bold{{f}}),
%\vspace{-1.5mm}
\label{eq:stereo_vel}
\end{equation} 
\noindent 
where $\mathbf{R}^b_c$ is a given extrinsic parameter between {the} IMU and {the} camera and ${^c}\tilde{\mathbf{v}}$ is the body velocity measurement obtained from a stereo camera. \textcolor{black}{Note that we can recover the depth of features thanks to a calibrated stereo camera. Unlike \cite{ci2016robust}, STEP does not establish the objective function to compute the body velocity quickly. Instead, the direct linear transformation (DLT)~\cite{andrew2001multiple} is exploited. For the same reason, the space position constraint defined in \cite{ci2016robust} is not adopted based on the mild assumption that the environment is almost static.}
\vspace{-2mm}

\subsection{Forward Kinematics Factor}\label{sec:fk}
The detailed derivation of the forward kinematics factor can be found in \cite{8hartley2018legged,hartley2017supplementary}. We adopt the forward kinematic measurement model as follows:%\vspace{-1mm}
\begin{equation}%\small
\begin{aligned}
{^b}\boldsymbol{\Gamma}_{R} &=\mathbf{R}^{\top} \boldsymbol{\Psi} \operatorname{Exp}\left(\delta {^b}\boldsymbol{\Gamma}_{R}\right) \\
{^b}\boldsymbol{\Gamma}_{p} &=\mathbf{R}^{\top}(\mathbf{s}-\mathbf{p})+\delta {^b}\boldsymbol{\Gamma}_{p},
\end{aligned}
%\vspace{-2mm}
\end{equation}
\noindent
where $\mathbf{R}$ is the body orientation; $\delta{^b}\boldsymbol{\Gamma}_{R}$ and $\delta{^b}\boldsymbol{\Gamma}_{p}$ represent small perturbations from ${^b}\boldsymbol{\Gamma}_{R}$ and ${^b}\boldsymbol{\Gamma}_{p}$, respectively \cite{8hartley2018legged, hartley2017supplementary}. The forward kinematics factor \textcolor{black}{$\mathbf{r}_{\mathcal{K}}(\mathcal{K}_{i},\mathcal{X}_k)=\big[\mathbf{r}_{\mathcal{K}_{R_{i}}}, \mathbf{r}_{\mathcal{K}_{p_{i}}}\big]$} is represented as follows:%\vspace{-2mm}
\begin{equation}%\small
\begin{aligned}
\mathbf{r}_{\mathcal{K}_{R_{i}}}\textcolor{black}{(\mathcal{K}_{i},\mathcal{X}_k)} &=\operatorname{Log} \left({^b}\boldsymbol{\Gamma}_{R,i}^{\top} \mathbf{R}_{i}^{\top} \boldsymbol{\Psi}_{i}\right) \\
\mathbf{r}_{\mathcal{K}_{p_{i}}}\textcolor{black}{(\mathcal{K}_{i},\mathcal{X}_k)} &=\mathbf{R}_{i}^{\top}\left(\mathbf{s}_{i}-\mathbf{p}_{i}\right)-{^b}\boldsymbol{\Gamma}_{p,i}.
\end{aligned}%\vspace{-2mm}
\end{equation}
%%%%%%%%%%%%%%%%%%%%%%%%%%%%%%%%%

\subsection{Foot Velocity Preintegration}\label{sec:foot preintegration}
In contrast to most previous studies, we do not {exploit} the {non-slip} assumption, which assumes the foot velocity is zero in the contact state. Instead, we take the information from the foot angular velocity ${^f}\boldsymbol\omega$ and linear velocity ${^f}{\boldsymbol{\nu}}$ expressed in the foot frame and associated noise{s} $\mathbf{n}^{\tilde{\boldsymbol\omega}}$ and $\mathbf{n}^{\tilde{\boldsymbol\nu}}$, respectively. Then, we preintegrate the information to find the change in {the} foot pose.

Similar to \cite{8hartley2018legged, forster2016manifold}, (\ref{eq:contPsi}) can be discretized based on the assumption that ${^f}\tilde{\boldsymbol\omega}$ is constant during sampling time $\Delta{t}$: \vspace{-2mm}
\begin{equation}
    \boldsymbol\Psi_j=\boldsymbol\Psi_i\prod_{k=i}^{j-1}\operatorname{Exp}({^f}\tilde{\boldsymbol\omega}_k-\mathbf{n}^{\tilde{\boldsymbol\omega}d}_k)\Delta{t},
    \vspace{-2mm}
\end{equation}
\noindent 
where $\mathbf{n}^{\tilde{\boldsymbol\omega}d}_{k}$ {is the discrete time noise represented in the foot frame} with covariance $\boldsymbol\Omega^{{\tilde{\boldsymbol\omega}}d}$ and {is} computed using sampling time $\Delta{t}$; $\boldsymbol\Omega^{{\tilde{\boldsymbol\omega}}d} = \frac{1}{\Delta{t}}\boldsymbol\Omega^{{\tilde{\boldsymbol\omega}}}$.

We {define} the preintegrated term $\Delta\boldsymbol\Psi_{ij}$ independent of the state as follows:\vspace{-3mm}
\begin{equation}
    \Delta\boldsymbol\Psi_{ij} \doteq \boldsymbol{\Psi}_i^{\top}\boldsymbol{\Psi}_j = \prod_{k=i}^{j-1}\operatorname{Exp}({^f}\tilde{\boldsymbol\omega}_k-\mathbf{n}^{\tilde{\boldsymbol\omega}d}_{k})\Delta{t}.
\label{eq:preintPsi}
\vspace{-2mm}
\end{equation}

Furthermore, we hope to isolate the noise from  {the} preintegrated measurement. Using (\ref{eq:ExpJac}) and (\ref{eq:commExp}), (\ref{eq:preintPsi}) can be approximated {as}:\vspace{-2mm}
\begin{equation}%\small
\begin{split}
        \Delta\boldsymbol\Psi_{ij} &\simeq \prod_{k=i}^{j-1}[\operatorname{Exp}({^f}\tilde{\boldsymbol\omega}_k\Delta{t})\operatorname{Exp}(-\mathbf{J}_r^k({^f}\tilde{\boldsymbol\omega}_k\Delta{t})\mathbf{n}^{\tilde{\boldsymbol\omega}d}_{k}\Delta{t})]\\
        &\doteq \Delta{\tilde{\boldsymbol\Psi}}_{ij}\operatorname{Exp}(-\delta\boldsymbol\psi_{ij}),\raisetag{20pt}
\end{split}\label{eq:approxPsi}
\vspace{-2mm}
\end{equation}
\noindent
where $\mathbf{J}^k_r({^f}\tilde{\boldsymbol\omega}_k\Delta{t})$ is the {right Jacobian} of $\mathrm{SO}(3)$ (refer to (\ref{eq:rightJac})){;} $\Delta{\tilde{\boldsymbol\Psi}}_{ij}\doteq\prod_{k=i}^{j-1}\operatorname{Exp}({^f}\tilde{\boldsymbol\omega}_k\Delta{t})$ {is} the {preintegrated foot orientation measurement} and its noise term {is} $\operatorname{Exp}(-\delta\boldsymbol\psi_{ij})\doteq \prod_{k=i}^{j-1}\operatorname{Exp}(-\Delta\tilde{\boldsymbol\Psi}_{k+1j}^{\top}\mathbf{J}_r^k({^f}\tilde{\boldsymbol\omega}_k\Delta{t})\mathbf{n}^{\tilde{\boldsymbol\omega}d}_{k}\Delta{t})$. The noise of $\Delta\tilde{\boldsymbol\Psi}_{ij}$ can be computed by taking {the} $\operatorname{Log}$ on both sides \cite{forster2016manifold} {as}:\vspace{-2mm}
\begin{equation}
    \delta\boldsymbol\psi_{ij} \simeq \sum_{k=i}^{j-1}\Delta\tilde{\boldsymbol\Psi}_{k+1j}^{\top}~\mathbf{J}^k_r({^f}\tilde{\boldsymbol\omega}_k\Delta{t})\mathbf{n}^{\tilde{\boldsymbol\omega}d}_{k}\Delta{t}.\vspace{-2mm}
\end{equation}

Similarly, by iteratively accumulating the changes in foot position obtained from foot velocity ${^f}\tilde{\boldsymbol{\nu}}$, the next foot position at image frame rate can be calculated {as:}\vspace{-2mm}
\begin{equation}
\label{eq:foot position calculation}
\mathbf{s}_j =\mathbf{s}_i+\sum_{k=i}^{j-1}[\boldsymbol{\Psi}_k({{{^f}\tilde{\boldsymbol{\nu}}}_k}-\mathbf{n}_k^{\tilde{\boldsymbol\nu}d})\Delta{t}],
\vspace{-3mm}
\end{equation}
\noindent where $\mathbf{n}^{\tilde{\boldsymbol\nu}d}_{k}$ {is the discrete time noise in the foot frame} with covariance $\boldsymbol\Omega^{\tilde{\boldsymbol\nu}d}$ and {is} computed using sampling time $\Delta{t}$; $\boldsymbol\Omega^{\tilde{\boldsymbol\nu}d} = \frac{1}{\Delta{t}}\boldsymbol\Omega^{\tilde{\boldsymbol\nu}}$.

Now, the foot position $\mathbf{s}_j$ at time $j$ is computed by adding the change in foot position to {the} previous foot position $\mathbf{s}_i$ at time $i$. Here, we assume that the body velocity used in (\ref{eq:foot velocity foot}) between $i$ and $j$ is constant.

To avoid dependency on foot position $\mathbf{s}_i$, the preintegated foot position measurement $\Delta\mathbf{s}_{ij}$, between $i$ and $j$ in the body frame, can be defined from (\ref{eq:foot position calculation}) as follows:\vspace{-2.5mm}
\begin{equation}%\small
\begin{split}
   \Delta\mathbf{s}_{ij}\doteq\boldsymbol{\Psi}_{i}^{\top}(\mathbf{s}_{j}-\mathbf{s}_{i}) &= \boldsymbol{\Psi}^{\top}_i \sum_{k=i}^{j-1}[\boldsymbol{\Psi}_{k}({{^f}\tilde{\boldsymbol{\nu}}_k}-\mathbf{n}_k^{\tilde{\boldsymbol\nu}d})\Delta t)] \\
   &=\sum_{k=i}^{j-1}[\Delta \boldsymbol{\Psi}_{ik}({{^f}\tilde{\boldsymbol{\nu}}_k}-\mathbf{n}_k^{\tilde{\boldsymbol\nu}d})\Delta t)].
\end{split}
\label{eq:notApprox}
\vspace{-4mm}
\end{equation}

Using~(\ref{eq:anticomm}) and (\ref{eq:Expassume}), (\ref{eq:notApprox}) can be approximated by ignoring high order noise terms as follows:\vspace{-2mm}
\begin{equation}%\small
\begin{split}
    \Delta \mathbf{s}_{ij} &\simeq \sum_{k=i}^{j-1}\left[\Delta \tilde{\boldsymbol\Psi}_{i k}\left(\mathbf{I}-(\delta \boldsymbol\psi_{i k})^{\wedge}\right)\left({{^f}\tilde{\boldsymbol{\nu}}_k}-\mathbf{n}_k^{\tilde{\boldsymbol\nu}d}\right)\Delta t\right]\\
    &\simeq \sum_{k=i}^{j-1}[\Delta\tilde{\boldsymbol\Psi}_{ik}{^f}\tilde{\boldsymbol\nu}_k\Delta{t}]\\
    &\quad - \sum_{k=i}^{j-1}[\Delta\tilde{\boldsymbol\Psi}_{ik}\mathbf{n}_k^{\tilde{\boldsymbol\nu}d}\Delta{t}- \Delta{\tilde{\boldsymbol\Psi}}_{ik}({^f}\tilde{\boldsymbol\nu}_k)^{\wedge}\delta\boldsymbol\psi_{ik}\Delta{t}] \\
    &\doteq \Delta\tilde{\mathbf{s}}_{ij} - \delta \mathbf{s}_{ij},
\label{}\vspace{-2mm}
\end{split}
\end{equation}%\normalsize
\noindent where we define the {preintegrated foot position measurement} {as} $\Delta \tilde{\mathbf{s}}_{i j}=\sum_{k=i}^{j-1}[\Delta\tilde{\boldsymbol\Psi}_{ik}{^f}\tilde{\boldsymbol\nu}_k\Delta{t}]$ and its noise {as} $\delta\mathbf{s}_{ij}=\sum_{k=i}^{j-1}[\Delta\tilde{\boldsymbol\Psi}_{ik}\mathbf{n}_k^{\tilde{\boldsymbol\nu}d}\Delta{t}- \Delta{\tilde{\boldsymbol\Psi}}_{ik}({^f}\tilde{\boldsymbol\nu}_k)^{\wedge}\delta\boldsymbol\psi_{ik}\Delta{t}]$.

Note that the bias update is not considered in this study because it contributes little to improving accuracy compared to its computational complexity.\vspace{-2mm}
\subsection{Preintegrated Foot Velocity Factor}\label{sec:cost function}
Finally, from Section \ref{sec:foot preintegration}, the preintegrated foot velocity residual \textcolor{black}{$\mathbf{r}_{\mathcal{V}}(\mathcal{V}_{ij},\mathcal{X}_k)=\big[\mathbf{r}_{\mathcal{V}_{R_{i}}}, \mathbf{r}_{\mathcal{V}_{p_{i}}}\big]$} can be defined:\vspace{-2mm}
\begin{equation}
\mathbf{r}_{\mathcal{V}_{R_{i}}}\textcolor{black}{(\mathcal{V}_{ij},\mathcal{X}_k)}=\operatorname{Log}(\Delta\tilde{\boldsymbol{\Psi}}_{i j}^{\top}\Delta{\boldsymbol{\Psi}}_{ij}),
\end{equation}
\begin{equation}\vspace{-2mm}
\mathbf{r}_{\mathcal{V}_{p_{i}}}\textcolor{black}{(\mathcal{V}_{ij},\mathcal{X}_k)}=\Delta{\mathbf{s}}_{i j} -\Delta \tilde{\mathbf{s}}_{i j}.
\end{equation}

The noise vector of {the} preintegrated measurement can be modeled as a zero-mean, {n}ormally distributed vector, $\delta\boldsymbol\eta_{ij}\doteq{[\delta{\boldsymbol\psi^{\top}_{ij}}, \delta\mathbf{s}_{ij}^{\top}]}^{\top}\sim\mathcal{N}(\mathbf{0}_{6\times1},\boldsymbol\Omega_{\boldsymbol\eta_{ij}})$. Similarly, the noise vector related to the sensor is denoted as $\delta\mathbf{n}_{j}\doteq{[{\mathbf{n}^{\tilde{\boldsymbol\omega}d}_{j}}^{\top},{\mathbf{n}^{\tilde{\boldsymbol\nu}d}_j}^{\top}]}^{\top}\sim\mathcal{N}(\mathbf{0}_{6\times1},\boldsymbol\Omega_{\mathbf{n}_{j}})$.

The noise propagation can be established in an iterative form {as follows}:
\begin{equation}
    \left[\begin{array}{c}
    \delta \boldsymbol\psi_{i j+1} \\
    \delta \mathbf{s}_{i j+1} \\
    \end{array}\right]= \mathbf{A}_j  
    \left[\begin{array}{c}
    \delta \boldsymbol\psi_{i j} \\
    \delta \mathbf{s}_{i j} \\
    \end{array}\right]+\mathbf{B}_j
    \left[\begin{array}{c}
    \mathbf{n}^{\tilde{\boldsymbol\omega}d}_{j} \\
    \mathbf{n}^{\tilde{\boldsymbol\nu}d}_j \\
    \end{array}\right],
\end{equation} 
\noindent where \footnotesize{
\begin{equation}\vspace{-1mm}
\begin{split}
\mathbf{A}_j=\left[\begin{array}{cc}
\mathbf{I}_{3\times3} & \mathbf{0}_{3\times3} \\
-\Delta \tilde{\boldsymbol\Psi}_{ij} ({^f}\tilde{\boldsymbol\nu}_j)^{\wedge} \Delta t & \mathbf{I}_{3\times3} \\
\end{array}\right],\\
\mathbf{B}_j=\left[\begin{array}{cc}
\mathbf{J}^j_r({^f}\tilde{\boldsymbol\omega}_j\Delta{t})     & \mathbf{0}_{3\times3} \\
\mathbf{0}_{3\times3}     & \Delta\tilde{\boldsymbol\Psi}_{ij}\Delta{t}
\end{array}\right]. \nonumber
\end{split}
\end{equation}} \normalsize

Thus, the preintegrated measurement covariance can be computed iteratively:
\vspace{-1mm}
\begin{equation} %\small 
\vspace{-1mm}
    \boldsymbol\Omega_{\boldsymbol\eta_{ij+1}} = \mathbf{A}_j \boldsymbol\Omega_{\boldsymbol\eta_{ij}} \mathbf{A}_j^{\top} + \mathbf{B}_j \boldsymbol\Omega_{\mathbf{n}_j} \mathbf{B}_j^{\top},\vspace{-1mm}
\end{equation} %\normalsize
\noindent
where $\boldsymbol\Omega_{\mathbf{n}_{j}}\in\mathbb{R}^{6\times6}$ is the covariance matrix with $\boldsymbol\Omega^{{\tilde{\boldsymbol\omega}}d}$ and $\boldsymbol\Omega^{\tilde{\boldsymbol\nu}d}$ as diagonal components as follows:\vspace{-1mm}
%\small
\begin{equation}
    \boldsymbol\Omega_{\mathbf{n}_j} = \left[\begin{array}{cc}
       \boldsymbol\Omega^{{\tilde{\boldsymbol\omega}}d}  & \mathbf{0}_{3\times3} \\
        \mathbf{0}_{3\times3} & \boldsymbol\Omega^{\tilde{\boldsymbol\nu}d}
    \end{array}\right].
\vspace{-1mm}
\end{equation}%\normalsize

Finally, it allows us to compute the preintegrated measurement covariance $\boldsymbol\Omega_{\boldsymbol\eta_{ij+1}}$ starting with $\boldsymbol\Omega_{\boldsymbol\eta_{ii}}=0$.

\section{Experimental Results}\label{sec:experimental results}
In this section, we demonstrate the performance of STEP in various conditions. We first tested STEP in the {G}azebo simulation with texture-less mountainous terrain and slippery area. Second, {we} evaluated {STEP} on a public dataset \cite{1camurri2020pronto} collected in a factory-like environment. Lastly, we conducted a challenging experiment in gravel environments with the Mini-Cheetah robot \cite{6katz2019mini}. 

To verify the performance of STEP, {we compared} the experiment{al} results with other state-of-the-art state estimators: (1) \textbf{Pronto} \cite{1camurri2020pronto}, which is an EKF-based algorithm using {an} IMU, leg odometry, and {a} camera in a loosely coupled manner{;} (2) \textcolor{black}{\textbf{VINS-Fusion} \cite{qin2019general}, a factor graph-based multi-sensor state estimator. In this comparison, a stereo camera and IMU configuration is used for fairness; and} (3) \textcolor{black}{\textbf{WALK-VIO} \cite{lim2021walk}{, which} is the previous version of STEP, utilizing} the leg kinematic constraint based on a non-slip assumption, is also compared. \textcolor{black}{To compare the effect of leg kinematic constraints only, the adaptive factor was not considered.}
% to prove the effect of {the} preintegrated foot velocity factor.

{We} implemented {the algorithms} on Ubuntu 18.04 with ROS Melodic. {We tested e}very experiment with an Intel Core i7-8700K CPU with 32GB of memory. \vspace{-2mm}

\subsection{Gazebo Simulation}
%\textit{1) Simulation Setup}

{We conducted t}he simulation using Gazebo, which closely interacts with ROS. Various quadruped robots could be considered, but we used ANYmal \cite{hutter2016anymal} as the target quadruped robot platform for simulation. For the sensor configurations, the stereo camera, Intel RealSense D435i\footnote{https://dev.intelrealsense.com/docs/stereo-depth-camera-d400}, and the IMU with a rate of 400Hz were used. In addition, we used the open-source {motion} controller Champ\footnote{https://github.com/chvmp/champ}.

\begin{figure}[t]
    \centering
    %\subfigure[]{\label{fig:anymalonslip}\includegraphics[width=0.41\linewidth]{figure/anymal_onslip.pdf}}
    {\label{fig:whiteworld}\includegraphics[width=1.0\linewidth]{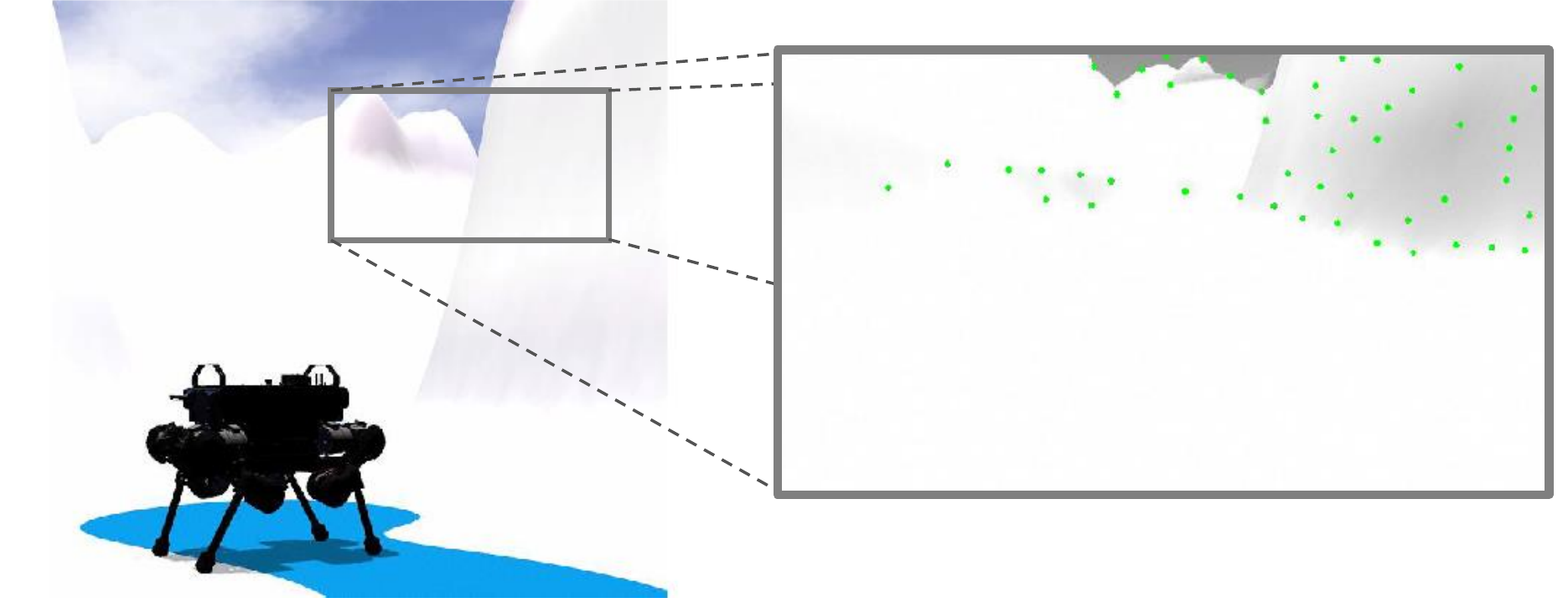}}
    \caption{ANYmal \cite{hutter2016anymal} on a slippery artifact in the simulation. The slippery surfaces are colored in blue. The right image in the gray box is obtained from the camera of ANYmal in white mountainous simulation environment in Gazebo. The {point} features tracked in VINS-Fusion \cite{qin2019general} are marked with green dots.}
    \label{fig:simulEnv}
%\vspace{-7mm}
\end{figure}

To prove the performance of STEP even in the extreme environments, we constructed the mountain terrain simulation world as shown in Fig. ~\ref{fig:simulEnv}. We designed the overall simulation environment to be bright and texture-less so that it {was} challenging for visual-dependent approaches. Moreover, {we included} the experiment{al} results on several slippery surfaces that {made} {non-slip} assumption invalid. The total distance traveled was approximately 200m.

{We} implemented and evaluated \textcolor{black}{several algorithms} (see Table~\ref{table:results}). As expected, due to the limitation of loosely coupled approaches, \textcolor{black}{Pronto \cite{1camurri2020pronto} degraded if any sensor data is not stable. Therefore, Pronto showed a relatively large error in the texture-less and slippery environment.} %Likewise, contact-aided IEKF \cite{hartley2020contact} deteriorated because of the slippery artifacts implemented in the simulation world.

\begin{figure*}[t]
    \centering
    \subfigure[][]{\label{fig:simulRes}\includegraphics[scale=0.42]{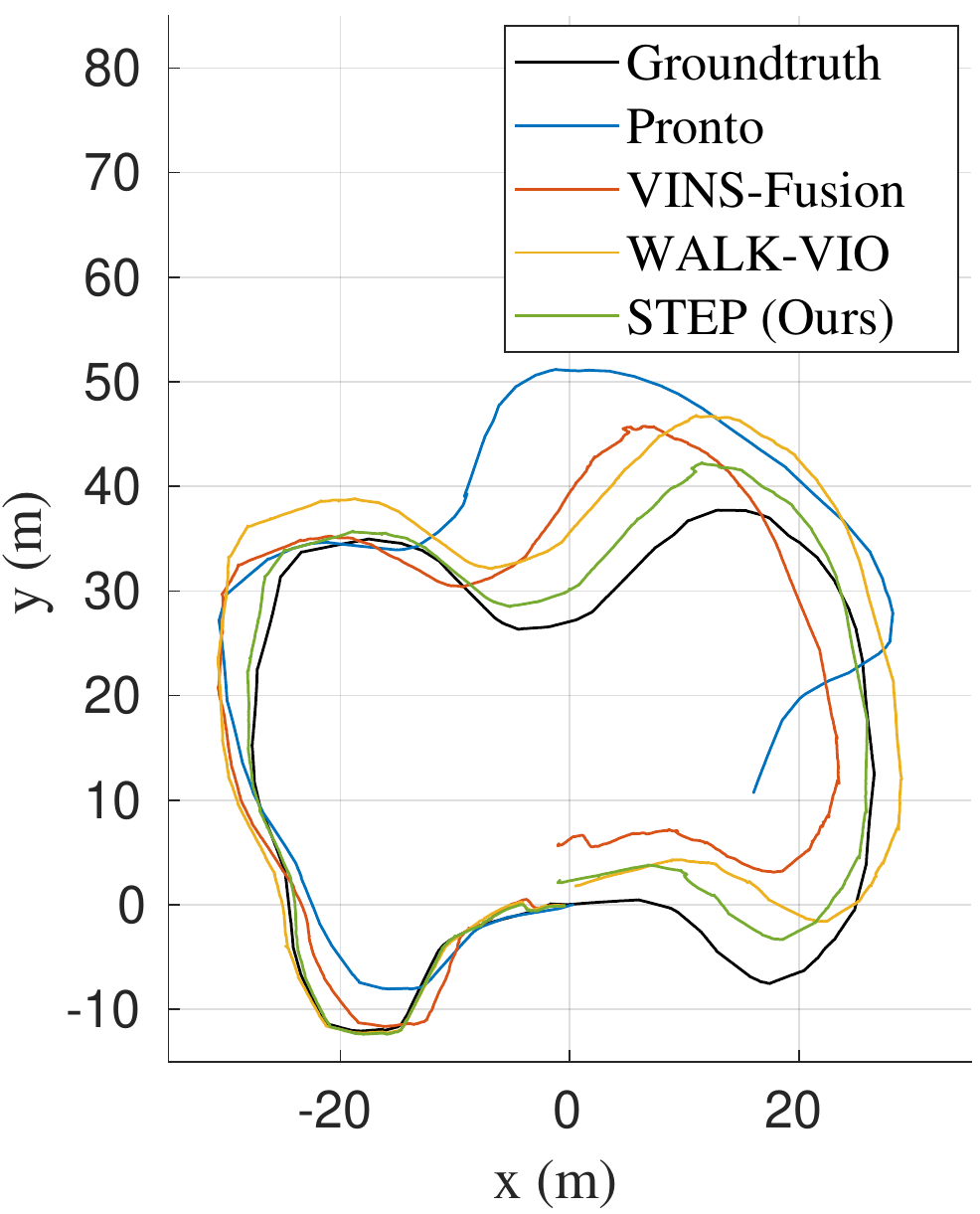}} \vspace{-1.5mm}
    \subfigure[][]{\label{fig:simul2Res}\includegraphics[scale=0.42]{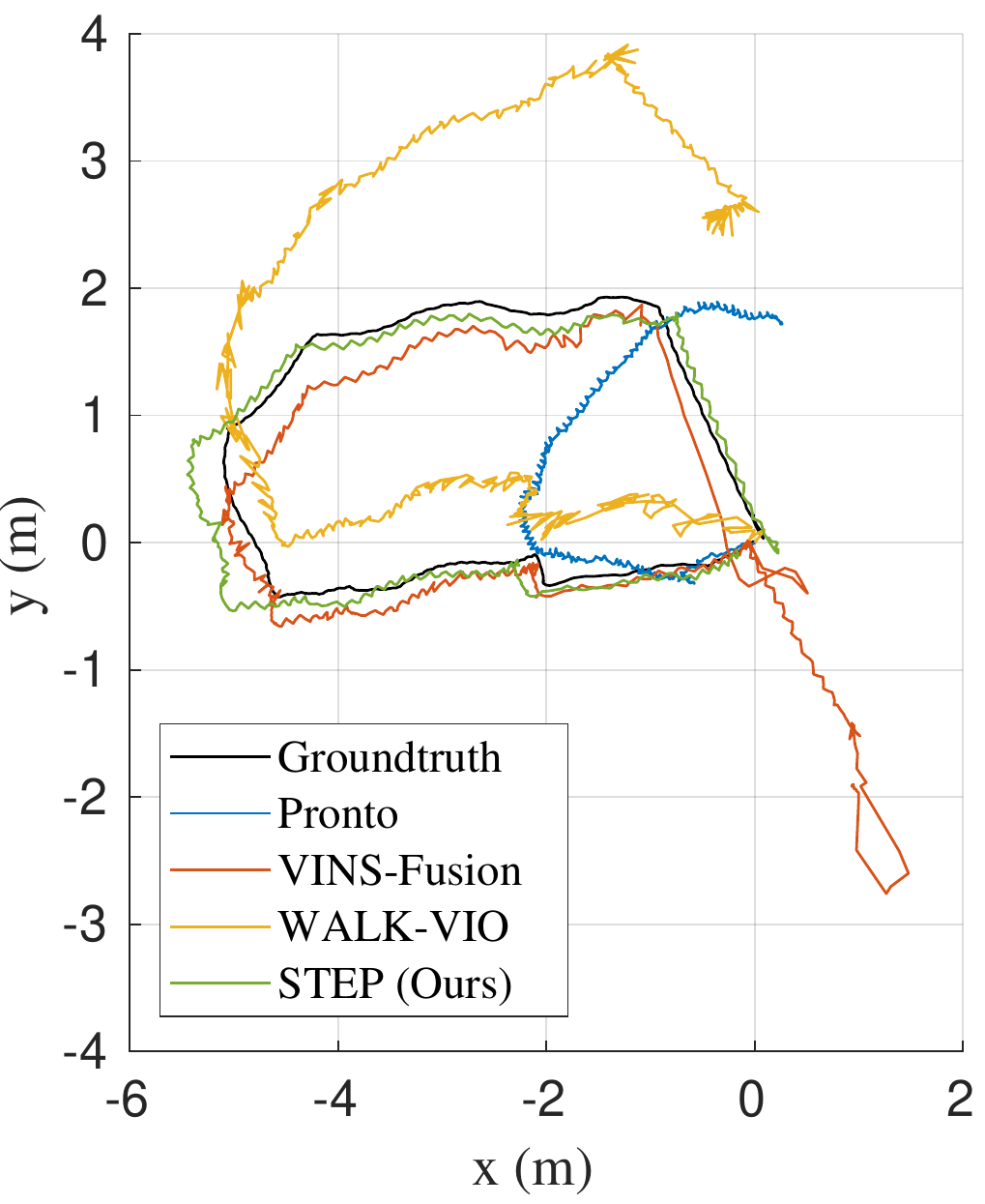}} 
    \subfigure[][]{\label{fig:FSCResMCRes}\includegraphics[scale=0.42]{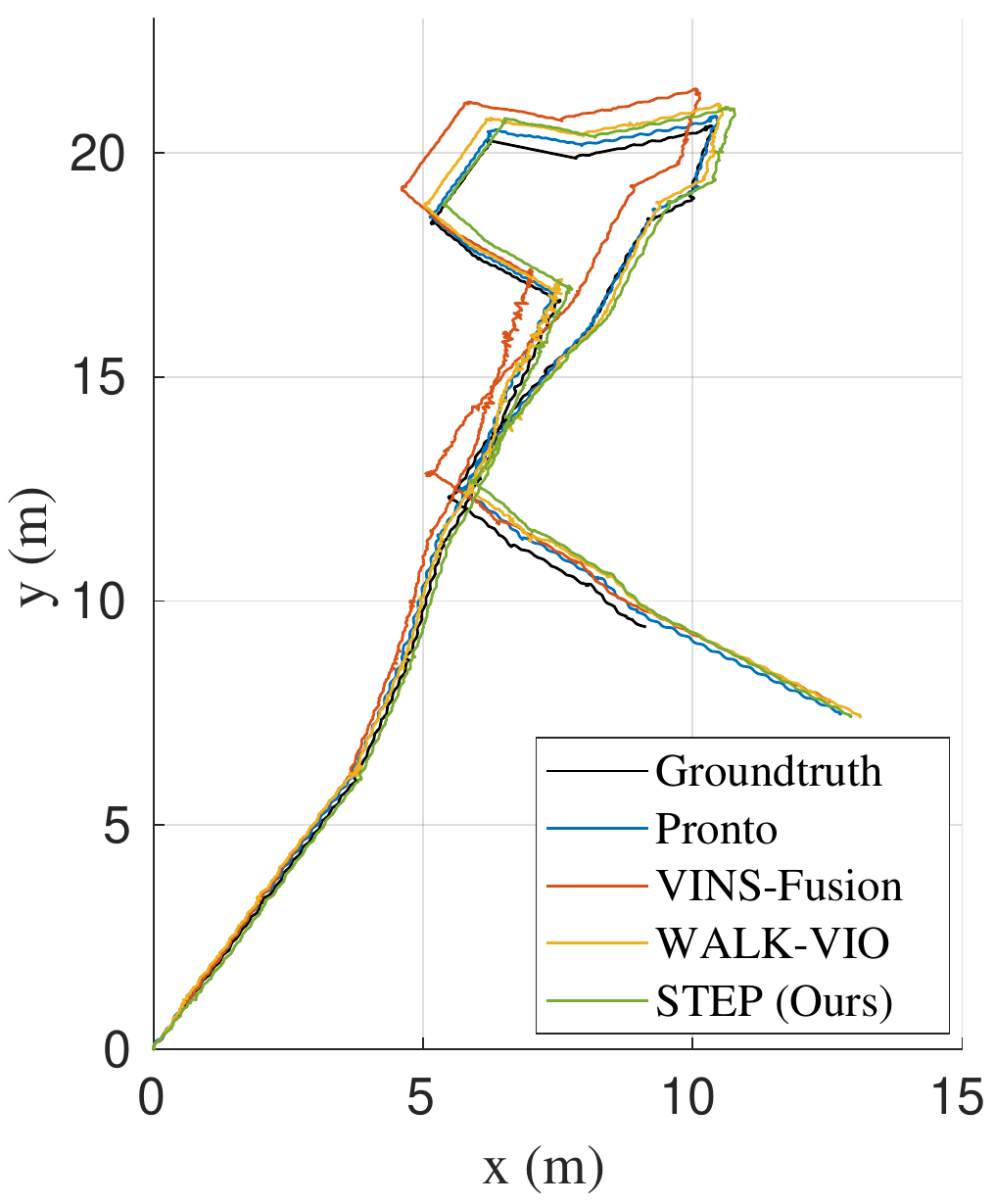}} 
    \subfigure[][]{\label{fig:MCRes}\includegraphics[scale=0.42]{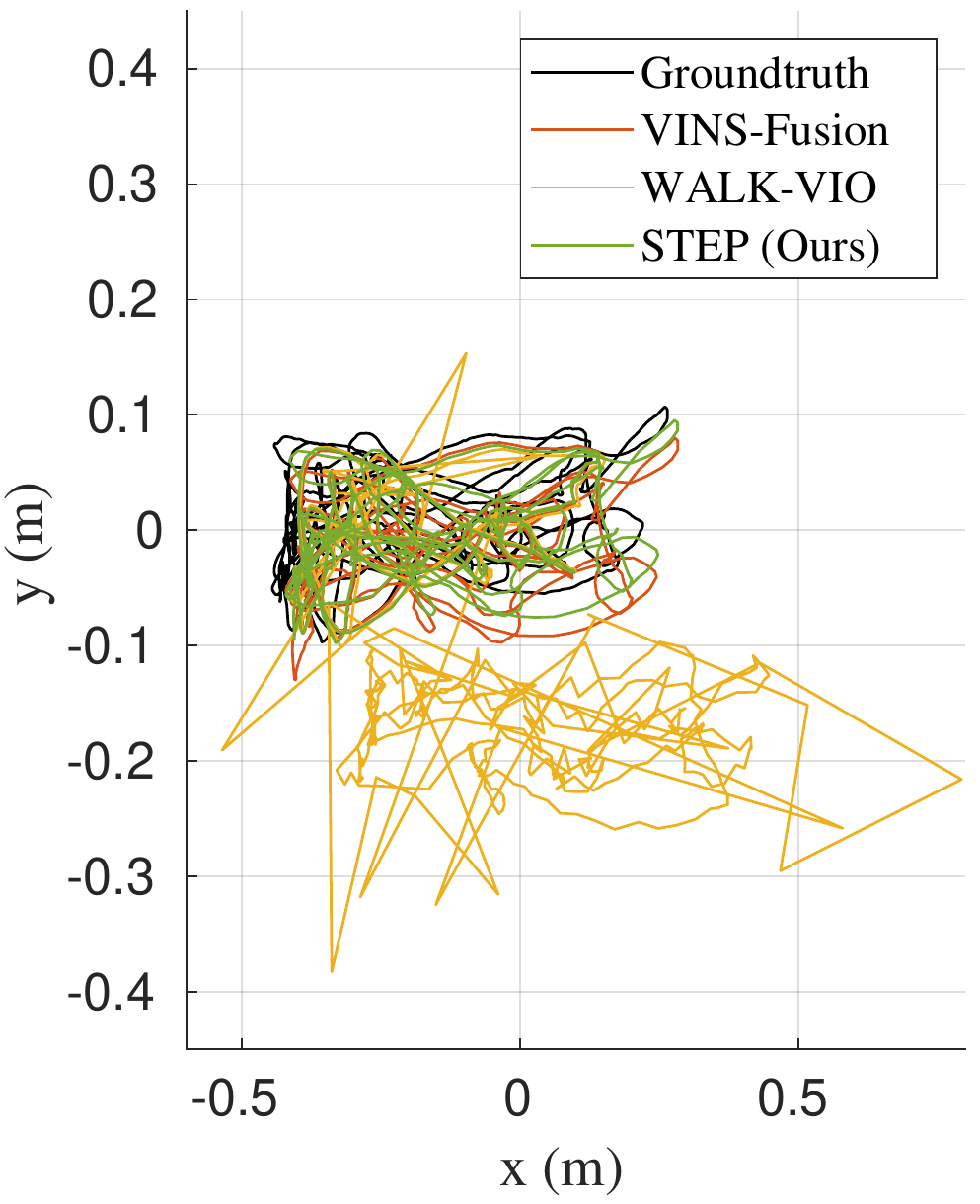}} 
    \caption{(a) The estimated trajectories of ANYmal in the simulation, especially {in} texture-less environment, and (b) in the slippery environment in the simulation, (c) in {the} FSC dataset, and (d) in the real-world experiment using Mini-Cheetah. \textcolor{black}{All starting points are zero.}} %Top view of trajectories estimated with Pronto (blue), VINS-Fusion (green), STEP with leg kinematics under the {non-slip} assumption (orange), and STEP (cyan). Ground truth is colored in magenta.}
    \label{fig:AllRes}
\vspace{-6mm}
\end{figure*}

Contrarily, the tightly coupled visual-inertial odometry system, VINS-Fusion \cite{qin2019general} showed relatively better performance. However, \textcolor{black}{WALK-VIO \cite{lim2021walk} degraded on the slippery terrain due to the {non-slip} assumption,} as described in Fig. ~\ref{fig:simul2Res}.

STEP outperformed others because it \textcolor{black}{estimates the foot pose more precisely because it never depends on a non-slip assumption. Note that the foot pose and the body pose are tightly coupled.} %fused visual, inertial, and kinematics information, even without a strong assumption. 
Fig.~\ref{fig:simulRes} and \ref{fig:simul2Res} illustrate the more reliable performance of STEP than the others in the simulation. %We marked the trajectories on slippery terrains in yellow circles.
\vspace{-1mm}

\subsection{Public Dataset Evaluation}
For legged robots, few public datasets are available. Thanks to the authors of \cite{1camurri2020pronto}, {we evaluated} STEP on the \textit{Fire Service College} (FSC) dataset\footnote{https://github.com/ori-drs/pronto}. The ANYmal \cite{hutter2016anymal} was utilized to acquire the FSC dataset. 
As described in Table~\ref{table:results} \textcolor{black}{and} Fig.~\ref{fig:FSCResMCRes}, the leg kinematics-aided algorithm shows slightly better performance. All algorithms presented adequate performance because this dataset does not include light changes or severe slip\textcolor{black}{pages}. This result also demonstrates that STEP can be considered an alternative to VINS-Fusion in the real world. \vspace{-5mm}

\subsection{Real-world Experiment} 
Mini-Cheetah \cite{6katz2019mini} has been validated to run dynamically and aggressively. Thus, we favored it as our target platform. {We conducted the experiment for approximately 2 minutes on gravel}. The main purpose of this experiment was to show that the approaches based on {non-slip} assumption deteriorate under severe slip{pages}. 
As described in Table~\ref{table:results} \textcolor{black}{and} Fig.~\ref{fig:MCRes}, STEP shows the best performance although we evaluated STEP on uneven and slippery terrain, and VINS-Fusion shows comparable achievement due to many abundant textures from the environment. As expected, \textcolor{black}{WALK-VIO} deteriorated. Since the actual robot includes more severe noise and modeling uncertainty, the performance of the state estimator has been more degraded than the simulation results. Note that Pronto was not evaluated \textcolor{black}{because} the effects of preintegrated velocity factor were only focused on. \vspace{-3mm}

\subsection{Discussion} \vspace{-1mm}
We show the effect of the preintegrated foot velocity factor by testing STEP in various environments. As can be seen in Table~\ref{table:results}, STEP showed a good performance even in the texture-less or slippery environments. We believe that this is because the preintegrated foot velocity factor helped improve foot pose estimation. Because the body pose is tightly coupled with the end effector pose, the body pose can also be estimated accurately. In contrast, WALK-VIO based on the non-slip assumption has a poor performance in the slippery environment. We conclude that it fails because adding only Gaussian noise could not compensate for the severe slippage effect, leading to inaccurately estimating the foot pose.\vspace{-2mm}
\section{Conclusion and Future Works}\label{sec:conclusion}
% \begin{figure}[t]
%     \centering
%     \includegraphics[width=0.7\linewidth]{figure/realworld_exp.png}
%     \caption{Mini-Cheetah \cite{6katz2019mini} on gravel. This experiment is designed to validate STEP {as} an alternative state estimator for legged robots in the general environment even under severe slip.}
%     \label{fig:real-world experiment}
% \vspace{-7mm}
% \end{figure}
This letter presents STEP, a novel method to deal with {state estimation of} legged robots in a general environment even under severe slip{pages}. The preintegrated foot velocity factor plays an essential role in accurate state estimation. The robustness of STEP was validated in the slippery, and texture-less environment{, in which the} {non-slip} assumption is prone to be violated. Moreover, STEP was demonstrated using a public dataset and {a} real-legged robot. The results show that STEP can be considered a {competitive} estimator for the legged robot. 

Further quantitative analysis of the preintegrated foot velocity factor, such as execution time and foot pose estimation accuracy{, should} be performed. In addition, a fusion of additional sensors, such as {the} LiDAR {sensor}, could be considered. \textcolor{black}{One assumption we used in (28) could be critical when the robot has aggressive motion between consecutive keyframes. Thus, a more robust way to measure the body velocity has to be developed.}
\vspace{-2mm}

\begin{table*}[t!]
%\begin{table}[t!]
%\tiny
\caption{{Translation ATE (Absolute trajectory error) and RPE (Relative pose error) over 1 $\mathrm{m}$ (RMSE, Unit: $\mathrm{m}$)}}\vspace{-2mm} %over 10~$\mathrm{m}$}\vspace{-2mm}
\label{table:results}
\centering
%\resizebox{16cm}{!}{
\resizebox{18cm}{!}{
\begin{tabular}{*{9}{c}}\toprule
\multirow{1}{*}{} & \multicolumn{4}{c}{\multirow{1}{*}{{Simulation (Gazebo)}}} 
                  & \multicolumn{2}{c}{\multirow{1}{*}{{Public dataset}}} 
                  & \multicolumn{2}{c}{\multirow{1}{*}{{Real robot platform}}}\\ \midrule
\multirow{2}{*}{} & \multicolumn{2}{c}{\multirow{1}{*}{{Texture-less environment}}} 
            & \multicolumn{2}{c}{\multirow{1}{*}{{Slippery environment}}} 
              & \multicolumn{2}{c}{\multirow{1}{*}{{Fire Service College (FSC)}}}
              & \multicolumn{2}{c}{\multirow{1}{*}{{Mini-Cheetah on gravel}}}\\
              {Traveled distance} & \multicolumn{2}{c}{$\approx 200\text{m}$} & \multicolumn{2}{c}{$\approx 15\text{m}$} & 
              \multicolumn{2}{c}{$\approx 60\text{m}$}  & 
              \multicolumn{2}{c}{$\approx 15\text{m}$} \\ \midrule
              {} & 
              \multicolumn{1}{c}{{ATE}} &
              \multicolumn{1}{c}{{RPE}} &
              \multicolumn{1}{c}{{ATE}} &
              \multicolumn{1}{c}{{RPE}} &
              \multicolumn{1}{c}{\scriptsize{ATE}} &
              \multicolumn{1}{c}{{RPE}} &
              \multicolumn{1}{c}{{ATE}} &
              \multicolumn{1}{c}{{RPE}} \\ \midrule
\multirow{1}{*}{Pronto\cite{1camurri2020pronto}} & 
\multirow{1}{*}{7.391} & \multirow{1}{*}{1.958} & \multirow{1}{*}{0.992} & \multirow{1}{*}{1.128
} & 
\multirow{1}{*}{$\textbf{0.462}$} & \multirow{1}{*}{$\textbf{1.427}$} & \multirow{1}{*}{N.A} &  \multirow{1}{*}{N.A} \\  
\midrule
\multirow{1}{*}{VINS-Fusion\cite{qin2019general}} & 
\multirow{1}{*}{1.882} & \multirow{1}{*}{1.459} & \multirow{1}{*}{0.766} & \multirow{1}{*}{1.052} & 
\multirow{1}{*}{{0.572}} & \multirow{1}{*}{{2.187}} & \multirow{1}{*}{0.091} &  \multirow{1}{*}{0.218} \\ 

\multirow{1}{*}{{WALK-VIO\cite{lim2021walk}}} & 
\multirow{1}{*}{1.701} & \multirow{1}{*}{1.454} & \multirow{1}{*}{0.682} & \multirow{1}{*}{1.188} &
\multirow{1}{*}{{0.518}} & \multirow{1}{*}{{1.867}} & \multirow{1}{*}{0.313} &  \multirow{1}{*}{0.461} \\ 

\rowcolor{tabGray}\multirow{1}{*}{{$\textbf{STEP (Ours)}$}} & 
\multirow{1}{*}{$\textbf{1.462}$} & \multirow{1}{*}{$\textbf{{1.355}}$} & \multirow{1}{*}{$\textbf{0.200}$} & \multirow{1}{*}{$\textbf{0.676}$} & 
\multirow{1}{*}{{{0.495}}} & \multirow{1}{*}{{{1.526}}} & \multirow{1}{*}{$\textbf{0.087}$} &  \multirow{1}{*}{$\textbf{0.156}$} \\
\bottomrule \vspace{-6mm}
\end{tabular}}
\end{table*}

\ifCLASSOPTIONcaptionsoff
  \newpage
\fi



\bibliographystyle{IEEEtran}

\bibliography{reference}

\begin{thebibliography}{10}
\providecommand{\url}[1]{#1}
\csname url@samestyle\endcsname
\providecommand{\newblock}{\relax}
\providecommand{\bibinfo}[2]{#2}
\providecommand{\BIBentrySTDinterwordspacing}{\spaceskip=0pt\relax}
\providecommand{\BIBentryALTinterwordstretchfactor}{4}
\providecommand{\BIBentryALTinterwordspacing}{\spaceskip=\fontdimen2\font plus
\BIBentryALTinterwordstretchfactor\fontdimen3\font minus
  \fontdimen4\font\relax}
\providecommand{\BIBforeignlanguage}[2]{{%
\expandafter\ifx\csname l@#1\endcsname\relax
\typeout{** WARNING: IEEEtran.bst: No hyphenation pattern has been}%
\typeout{** loaded for the language `#1'. Using the pattern for}%
\typeout{** the default language instead.}%
\else
\language=\csname l@#1\endcsname
\fi
#2}}
\providecommand{\BIBdecl}{\relax}
\BIBdecl

\bibitem{3bloesch2013state}
M.~Bloesch, C.~Gehring, P.~Fankhauser, M.~Hutter, M.~A. Hoepflinger, and
  R.~Siegwart, ``State estimation for legged robots on unstable and slippery
  terrain,'' in \emph{Proc. IEEE/RSJ International Conference on Intelligent
  Robots and Systems (IROS)}, 2013, pp. 6058--6064.

\bibitem{6katz2019mini}
B.~Katz, J.~Di~Carlo, and S.~Kim, ``Mini cheetah: A platform for pushing the
  limits of dynamic quadruped control,'' in \emph{Proc. IEEE International
  Conference on Robotics and Automation (ICRA)}, 2019, pp. 6295--6301.

\bibitem{hartley2020contact}
R.~Hartley, M.~Ghaffari, R.~M. Eustice, and J.~W. Grizzle, ``Contact-aided
  invariant extended kalman filtering for robot state estimation,'' \emph{The
  International Journal of Robotics Research}, vol.~39, no.~4, pp. 402--430,
  2020.

\bibitem{11teng2021legged}
S.~Teng, M.~W. Mueller, and K.~Sreenath, ``Legged robot state estimation in
  slippery environments using invariant extended {K}alman filter with velocity
  update,'' \emph{arXiv preprint arXiv:2104.04238}, 2021.

\bibitem{10fourmy2021contact}
M.~Fourmy, T.~Flayols, N.~Mansard, and J.~Sol{\`a}, ``Contact forces
  pre-integration for the whole body estimation of legged robots,'' in
  \emph{Proc. IEEE International Conference on Robotics and Automation (ICRA)},
  2021.

\bibitem{14kim2021legged}
J.-H. Kim, S.~Hong, G.~Ji, S.~Jeon, J.~Hwangbo, J.-H. Oh, and H.-W. Park,
  ``Legged robot state estimation with dynamic contact event information,''
  \emph{IEEE Robotics and Automation Letters}, vol.~6, no.~4, pp. 6733--6740,
  2021.

\bibitem{8hartley2018legged}
R.~Hartley, J.~Mangelson, L.~Gan, M.~G. Jadidi, J.~M. Walls, R.~M. Eustice, and
  J.~W. Grizzle, ``Legged robot state-estimation through combined forward
  kinematic and preintegrated contact factors,'' in \emph{Proc. IEEE
  International Conference on Robotics and Automation (ICRA)}, 2018, pp.
  4422--4429.

\bibitem{9hartley2018hybrid}
R.~Hartley, M.~G. Jadidi, L.~Gan, J.-K. Huang, J.~W. Grizzle, and R.~M.
  Eustice, ``Hybrid contact preintegration for visual-inertial-contact state
  estimation using factor graphs,'' in \emph{Proc. IEEE/RSJ International
  Conference on Intelligent Robots and Systems (IROS)}, 2018, pp. 3783--3790.

\bibitem{barrau2015non}
A.~Barrau, ``Non-linear state error based extended kalman filters with
  applications to navigation,'' Ph.D. dissertation, Mines Paristech, 2015.

\bibitem{forster2016manifold}
C.~Forster, L.~Carlone, F.~Dellaert, and D.~Scaramuzza, ``On-manifold
  preintegration for real-time visual--inertial odometry,'' \emph{IEEE
  Transactions on Robotics}, vol.~33, no.~1, pp. 1--21, 2016.

\bibitem{1camurri2020pronto}
M.~Camurri, M.~Ramezani, S.~Nobili, and M.~Fallon, ``Pronto: A multi-sensor
  state estimator for legged robots in real-world scenarios,'' \emph{Frontiers
  in Robotics and AI}, vol.~7, p.~68, 2020.

\bibitem{13wisth2019robust}
D.~Wisth, M.~Camurri, and M.~Fallon, ``Robust legged robot state estimation
  using factor graph optimization,'' \emph{IEEE Robotics and Automation
  Letters}, vol.~4, no.~4, pp. 4507--4514, 2019.

\bibitem{2wisth2020preintegrated}
------, ``Preintegrated velocity bias estimation to overcome contact
  nonlinearities in legged robot odometry,'' in \emph{Proc. IEEE International
  Conference on Robotics and Automation (ICRA)}, 2020, pp. 392--398.

\bibitem{15wisth2021vilens}
------, ``{VILENS: Visual, inertial, lidar, and leg odometry for all-terrain
  legged robots},'' \emph{arXiv preprint arXiv:2107.07243}, 2021\color{black}.

\bibitem{lim2021walk}
H.~Lim, B.~Yu, Y.~Kim, J.~Byun, S.~Kwon, H.~Park, and H.~Myung, ``{WALK-VIO}:
  Walking-motion-adaptive leg kinematic constraint visual-inertial odometry for
  quadruped robots,'' \emph{arXiv preprint arXiv:2111.15164},
  2021\color{black}.

\bibitem{5hwangbo2016probabilistic}
J.~Hwangbo, C.~D. Bellicoso, P.~Fankhauser, and M.~Hutter, ``Probabilistic foot
  contact estimation by fusing information from dynamics and
  differential/forward kinematics,'' in \emph{Proc. IEEE/RSJ International
  Conference on Intelligent Robots and Systems (IROS)}, 2016, pp. 3872--3878.

\bibitem{12camurri2017probabilistic}
M.~Camurri, M.~Fallon, S.~Bazeille, A.~Radulescu, V.~Barasuol, D.~G. Caldwell,
  and C.~Semini, ``Probabilistic contact estimation and impact detection for
  state estimation of quadruped robots,'' \emph{IEEE Robotics and Automation
  Letters}, vol.~2, no.~2, pp. 1023--1030, 2017.

\bibitem{4jenelten2019dynamic}
F.~Jenelten, J.~Hwangbo, F.~Tresoldi, C.~D. Bellicoso, and M.~Hutter, ``Dynamic
  locomotion on slippery ground,'' \emph{IEEE Robotics and Automation Letters},
  vol.~4, no.~4, pp. 4170--4176, 2019.

\bibitem{hutter2016anymal}
M.~Hutter, C.~Gehring, D.~Jud, A.~Lauber, C.~D. Bellicoso, V.~Tsounis,
  J.~Hwangbo, K.~Bodie, P.~Fankhauser, M.~Bloesch \emph{et~al.}, ``Anymal-a
  highly mobile and dynamic quadrupedal robot,'' in \emph{Proc. IEEE/RSJ
  International Conference on Intelligent Robots and Systems (IROS)}, 2016, pp.
  38--44.

\bibitem{LGchirikjian2009stochastic}
G.~S. Chirikjian, \emph{Stochastic Models, Information Theory, and Lie Groups,
  Volume 2: Analytic Methods and Modern Applications}.\hskip 1em plus 0.5em
  minus 0.4em\relax Springer Science \& Business Media, 2009, vol.~2.

\bibitem{LGabsil2009optimization}
P.-A. Absil, R.~Mahony, and R.~Sepulchre, \emph{Optimization Algorithms on
  Matrix Manifolds}.\hskip 1em plus 0.5em minus 0.4em\relax Princeton
  University Press, 2009.

\bibitem{LGhall2015lie}
B.~Hall, \emph{Lie Groups, Lie Algebras, and Representations: an Elementary
  Introduction}.\hskip 1em plus 0.5em minus 0.4em\relax Springer, 2015, vol.
  222.

\bibitem{barfoot2014associating}
T.~D. Barfoot and P.~T. Furgale, ``Associating uncertainty with
  three-dimensional poses for use in estimation problems,'' \emph{IEEE
  Transactions on Robotics}, vol.~30, no.~3, pp. 679--693, 2014.

\bibitem{qin2019general}
T.~Qin, J.~Pan, S.~Cao, and S.~Shen, ``A general optimization-based framework
  for local odometry estimation with multiple sensors,'' \emph{arXiv preprint
  arXiv:1901.03638}, 2019.

\bibitem{huber1992robust}
P.~J. Huber, ``Robust estimation of a location parameter,'' in
  \emph{Breakthroughs in Statistics}.\hskip 1em plus 0.5em minus 0.4em\relax
  Springer, 1992, pp. 492--518\color{black}.

\bibitem{more1978levenberg}
J.~J. Mor{\'e}, ``{The Levenberg-Marquardt algorithm}: implementation and
  theory,'' \emph{Numerical Analysis}, pp. 105--116, 1978.

\bibitem{ci2016robust}
W.~Ci and Y.~Huang, ``\color{black}{A} robust method for ego-motion estimation
  in urban environment using stereo camera,'' \emph{Sensors}, vol.~16, no.~10,
  p. 1704, 2016.

\bibitem{andrew2001multiple}
A.~M. Andrew, ``Multiple view geometry in computer vision,'' \emph{Kybernetes},
  2001\color{black}.

\bibitem{hartley2017supplementary}
R.~Hartley, J.~Mangelson, L.~Gan, M.~G. Jadidi, J.~M. Walls, R.~M. Eustice, and
  J.~W. Grizzle, ``Supplementary material: legged robot state-estimation
  through combined kinematic and preintegrated contact factors,''
  \emph{University of Michigan, Techsensoff. Rep., Feb}, 2017.

\end{thebibliography}
\end{document}